\documentclass[letterpaper, 10 pt, conference]{ieeeconf}
\IEEEoverridecommandlockouts
\let\labelindent\relax
\usepackage{graphicx} \graphicspath{ {figures/} }
\usepackage{amsmath,amssymb,mathabx}
\usepackage[cal=cm]{mathalfa}
\usepackage{mathtools}
\usepackage{algorithmic}
\usepackage[ruled]{algorithm2e}
\usepackage{acronym}
\usepackage{enumitem}
\usepackage{booktabs}
\usepackage{hyperref}
\usepackage{balance}
\usepackage{xspace,setspace}
\usepackage[skip=3pt,font=small]{subcaption}
\usepackage[skip=3pt,font=small]{caption}
\usepackage[misc]{ifsym}
\usepackage[dvipsnames]{xcolor}
\usepackage[capitalise]{cleveref}
\usepackage{tabularx,multirow,array,makecell}
\usepackage{overpic}
\usepackage{cite}

\makeatletter
\DeclareRobustCommand\onedot{\futurelet\@let@token\@onedot}
\def\@onedot{\ifx\@let@token.\else.\null\fi\xspace}
\def\eg{\emph{e.g}\onedot} 
\def\ie{\emph{i.e}\onedot} 
 
\def\etc{\emph{etc}\onedot}

\makeatother

\crefname{algocf}{alg.}{algs.}
\Crefname{algocf}{Algorithm}{Algorithms}

\DeclareMathOperator*{\kl}{KL}

\makeatletter
\def\BState{\State\hskip-\ALG@thistlm}
\makeatother

\makeatletter
\renewcommand{\paragraph}{%
  \@startsection{paragraph}{4}%
  {\z@}{0ex \@plus 0ex \@minus 0ex}{-1em}%
  {\hskip\parindent\normalfont\normalsize\bfseries}%
}
\makeatother

\crefname{algocf}{alg.}{algs.}
\Crefname{algocf}{Algorithm}{Algorithms}

\definecolor{gblue}{HTML}{4285F4}
\definecolor{gred}{HTML}{DB4437}

\acrodef{gta}[GTA V]{Grand Theft Auto V}
\acrodef{sf}[SF]{Social Force}
\acrodef{ioc}[IOC]{Inverse Optimal Control}
\acrodef{irl}[IRL]{Inverse Reinforcement Learning}
\acrodef{gnn}[GNNs]{Graph Neural Networks}
\acrodef{rnn}[RNN]{Recurrent Neural Networks}
\acrodef{gmm}[GMM]{Gaussian Mixture Model}
\acrodef{vae}[VAE]{Variational Auto-Encoder}
\acrodef{gat}[GAT]{Graph Attention Network}
\acrodef{gcn}[GCN]{Graph Convolution Network}
\acrodef{gail}[GAIL]{Generative Adversarial Imitation Learning}
\acrodef{kl}[KL]{Kullback-Leibler}

\title{\LARGE \bf \fontsize{15.5}{15.5}\selectfont
Congestion-aware Multi-agent Trajectory Prediction for Collision Avoidance
}

\author{Xu Xie$^{1}$\quad{}Chi Zhang$^{1}$\quad{}Yixin Zhu$^{1}$\quad{}Ying Nian Wu$^{1}$\quad{}Song-Chun Zhu$^{1}$% <-this % stops a space
\thanks{UCLA Center for Vision, Cognition, Learning, and Autonomy (VCLA) at Statistics Department. Emails: {\tt\small \{xiexu, chi.zhang, yixin.zhu\}@ucla.edu}, \tt\{ywu, sczhu\}@stat.ucla.edu.}% <-this % stops a space
\thanks{The work reported herein was supported by ONR N00014-19-1-2153, ONR MURI N00014-16-1-2007, and DARPA XAI N66001-17-2-4029.}%
}

\begin{document}

\maketitle
\thispagestyle{empty}
\pagestyle{empty}

\begin{abstract}
Predicting agents' future trajectories plays a crucial role in modern AI systems, yet it is challenging due to intricate interactions exhibited in multi-agent systems, especially when it comes to collision avoidance. To address this challenge, we propose to learn congestion patterns as contextual cues explicitly and devise a novel ``Sense--Learn--Reason--Predict'' framework by exploiting advantages of three different doctrines of thought, which yields the following desirable benefits: (i) Representing congestion as contextual cues via latent factors subsumes the concept of social force commonly used in physics-based approaches and implicitly encodes the distance as a cost, similar to the way a planning-based method models the environment. (ii) By decomposing the learning phases into two stages, a ``student'' can learn contextual cues from a ``teacher'' while generating collision-free trajectories. To make the framework computationally tractable, we formulate it as an optimization problem and derive an upper bound by leveraging the variational parametrization. In experiments, we demonstrate that the proposed model is able to generate collision-free trajectory predictions in a synthetic dataset designed for collision avoidance evaluation and remains competitive on the commonly used NGSIM US-101 highway dataset. Source code and dataset tools can be accessed via \href{https://github.com/xuxie1031/CollisionFreeMultiAgentTrajectoryPrediciton}{\color{blue}{Github}}.
\end{abstract}

\setstretch{0.96}

\section{Introduction}\label{sec:intro}

Since its inception, perceiving~\cite{heider1944experimental} and understanding~\cite{johansson1973visual} motions has become a key indicator for an intelligent system to interact with other agents in the environment felicitously. Unlike other topics (\eg, action understanding, or activity analysis), trajectory prediction is unique and proves to be challenging as it requires inference about \emph{multiple} agents in the \emph{future} yet to be observed. In literature, trajectory prediction can be roughly categorized into three directions~\cite{rudenko2019human}.

Modern modeling approaches first adopt a \textbf{physics-based} fashion~\cite{zhu1991hidden} in a ``Sense--Predict'' framework~\cite{rudenko2019human} by directly forward simulating pre-defined and explicit dynamic models based on Newton's laws of motion. Although physics-based cues are robust prior knowledge, this family of models tends to be too brittle to handle noisy real-world data, especially in multi-agent scenarios where different agents may possess various types of dynamic models. Recent multi-model approaches~\cite{li2005survey} attempt to alleviate these difficulties.

In parallel, \textbf{pattern-based} approaches~\cite{tadokoro1993stochastic} tackle the trajectory prediction problem by learning different function approximators directly from data, following a ``Sense--Learn--Predict'' paradigm~\cite{rudenko2019human}. The essence of this stream of work is to leverage the power of data to provide a data-driven account of the solution, which has received increasing attention over the past few years due to readily available large datasets. However, such methods naturally suffer from interpretability issues and tend to overfit with a large space of parameters.

By taking a teleological stand and assuming rational agents, \textbf{planning-based} approaches~\cite{bruce2004better} model the trajectory prediction problem by minimizing various costs, either by forward planning or inverse optimal control, following a ``Sense--Reason--Act'' principle~\cite{rudenko2019human}. However, such a \emph{normative} perspective, which models what agents \emph{ought} to do, may differ from real-world scenarios as the decision-making process often deviates from extreme rationality~\cite{stein1996without,kahneman1973psychology}.

Although different doctrines of thought have mostly been developing independently in literature, we argue that they do not conflict with each other and seek to answer how we can possibly fuse them and take advantage of these approaches to construct a new ``Sense--Learn--Reason--Predict'' framework. To give a desirable solution to this question, in this work, we start by answering the following three questions: (i) By reducing the parameter space, can a proper intermediate representation help to inject a better inductive bias for \textbf{pattern-based} approaches? Can such a representation be more generic and easy to, either explicitly or implicitly, incorporate the rational agent assumption in \textbf{planning-based} approaches and the physical constraints in \textbf{physics-based} approaches? (ii) Instead of using a single-stage learning process, will a well-designed multi-stage learning process improve the performance? (iii) Can such a design help emerge some crucial characteristics in multi-agent trajectory prediction, \eg, collision avoidance?

Specifically, we address the challenging problems in the task of \emph{collision-free multi-agent} trajectory prediction. In literature, first-order pattern-based approaches directly regress the trajectory based on the training data by fitting the position-based local transition patterns by either discrete cell~\cite{kruse1998camera,thompson2009probabilistic,kucner2013conditional,ballan2016knowledge,molina2018modelling}, continuous position~\cite{joseph2011bayesian,ferguson2015real,kucner2017enabling}, or graph-based representations~\cite{liao2003voronoi,vasquez2009incremental,chen2016augmented}, without any semantics-based intermediate representation (except human body motions~\cite{quintero2014pedestrian,minguez2018pedestrian}). Although higher-order pattern-based approaches incorporate some sorts of context, mostly in terms of relations between objects~\cite{kim2017probabilistic,altche2017lstm,park2018sequence,ding2019predicting,deo2018multi,deo2018convolutional,dai2019modeling,li2019coordination,tang2019multiple,zhao2019multi,srikanth2019infer}, they possess limited capability to emerge collision-free trajectories or verify whether the learning process or the learned model can do so. In contrast, we propose a learning method that incorporates high-level context cues of \emph{congestion}, aiming at emerging collision-free trajectories and qualitatively verifying them.

\begin{figure*}[t!]
    \centering
    \includegraphics[width=0.98\linewidth]{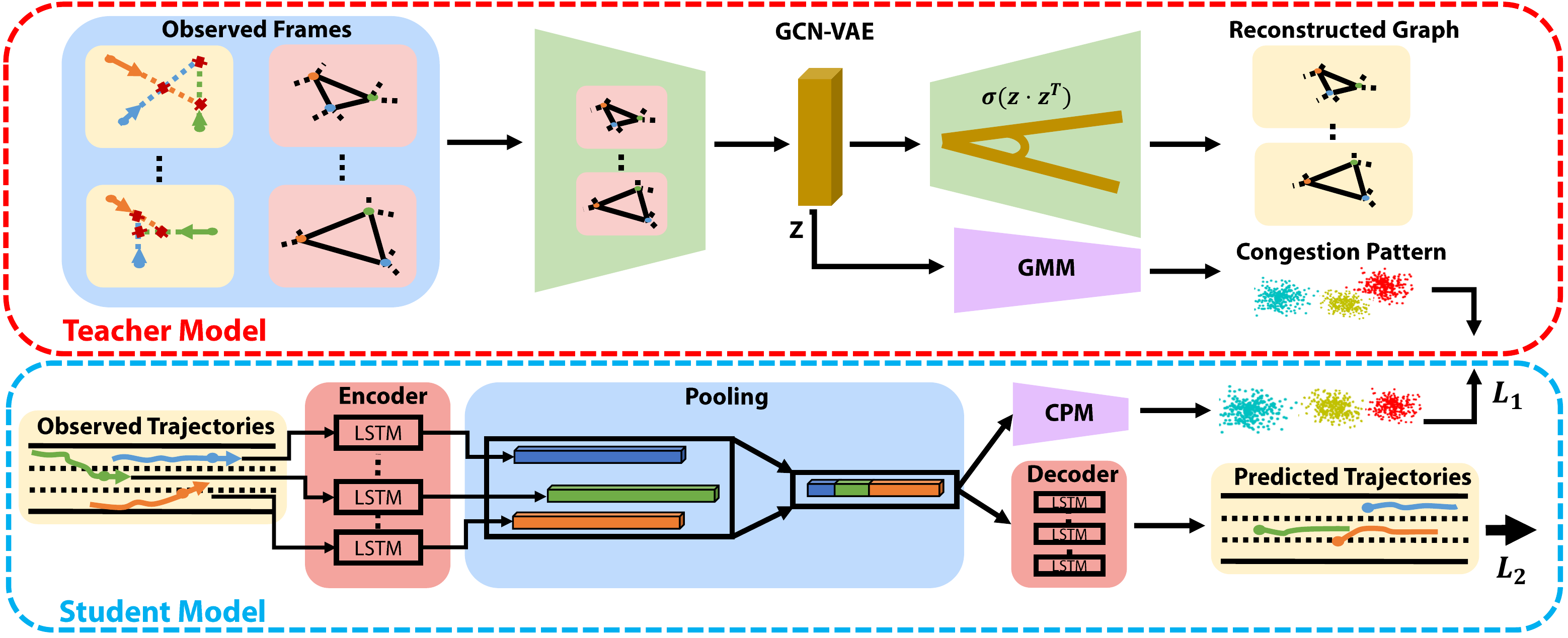}
    \caption{\textbf{The proposed architecture for congestion-aware multi-agent trajectory prediction.} The teacher model (top) is composed of the frame-wise graph construction module, and the \ac{gcn}-VAE graph encoder and decoder. The learned latents are passed to a \acs{gmm} and used to unsupervisedly learn the multi-modal congestion patterns. The student model (bottom) makes prediction based on the observed trajectories. It follows the encoder-pooling-decoder design and uses the CPM module to match the teacher's congestion patterns. The loss terms $L_1$ and $L_2$ are defined in \cref{equ:KL_new} and \cref{equ:traj}, respectively.}
    \label{fig:architecture}
\end{figure*}

The proposed method offers three unique advantages over prior methods. 
First, we represent the contextual cues by congestion via graph-based generative learning, wherein the node of the graph is the agent, and the edge of the graph is a measurement of the distance between two agents. Such a representation subsumes the \acf{sf} commonly used in physics-based approaches; it not only models pair-wise \ac{sf} but also provides a holistic view through graph modeling. Moreover, it implicitly encodes the (relative) distance as the cost, similar to how a planning-based method models the environment. Specifically, we use a \ac{gmm} to summarize the congestion patterns to (i) properly accounted for various modes in the congestion patterns, and (ii) reduce parameter space for the training process while maintaining a high task performance.

Second, we decouple the ``Sense--Learn--Reason--Predict'' framework into two processes: (i) A teacher ``senses'' and ``learns'' the contextual patterns (knowledge) as in a pure pattern-based approach. (ii) Instead of directly learning from observational data, a student reconstructs and ``reasons'' about the knowledge by minimizing a cost compared to what the teacher ``learns'', similar to a planning-based approach, while simultaneously ``predicting'' the future trajectory. Note that although such a design may look similar to GAN-based models for pedestrian trajectory predictions~\cite{gupta2018social,sadeghian2019sophie,kosaraju2019social}, they differ fundamentally. In GAN models, both the discriminator and the generator only focus on the predicted trajectory without explicit context modeling. In contrast, here in the proposed method, the student generates a trajectory supervised by explicitly learned contextual cues provided by the teacher. Such a design enables the student to learn from an inductive bias provided by the teacher, resulting in a faster training process and additional contextual constraints on the generated trajectories instead of random sampling.

Third, we formulate an optimization problem to bridge the connection between the congestion patterns and the learning objective of collision-free trajectory prediction, wherein the trajectories generated by the model are constrained by a learned congestion pattern distribution. By leveraging the variational parametrization, we derive an upper bound to make this optimization problem computationally tractable. 

The final model recruits an encoder--pooling--decoder network design and is therefore compatible with many existing trajectory prediction architectures (\eg, \cite{deo2018convolutional,alahi2016social}). In the experiments, we show the superiority of the proposed method in collision-free trajectory prediction in a new synthetic dataset designed to evaluate collision avoidance. Furthermore, the model remains competitive on the typical benchmark of the NGSIM US-101 highway dataset~\cite{colyar2007us}.

\section{Related Works}\label{sec:related_works}

\paragraph*{Contextual Cues of Moving Agents}

Agents' decisions on their motions depend on other agents' behaviors and interactions. In literature, such contextual cues are traditionally modeled by \ac{sf}~\cite{helbing1995social,luber2010people,elfring2014learning,ferrer2014behavior,kretzschmar2014learning,robicquet2016learning,karasev2016intent,xie2017learning,zhu2020dark} or integrated into motion policies or cost functions~\cite{wu2018probabilistic,zechel2019pedestrian,muench2019composable}. Recent data-driven approaches, if ever, only implicitly learn the contextual cues by training on large datasets~\cite{kim2017probabilistic,altche2017lstm,park2018sequence,ding2019predicting,deo2018multi,deo2018convolutional,dai2019modeling,li2019coordination,tang2019multiple,srikanth2019infer,kuhnt2016understanding,cui2019multimodal,hong2019rules}. In contrast, we propose to explicitly use congestion as the contextual cues, which accounts for \ac{sf} and is compatible with modern learning methods in pattern-based approaches.

\paragraph*{Congestion Detection and Congestion Pattern}

The notion of congestion has proven to be useful in various applications~\cite{bauza2010road,lin2011road,milojevic2014distributed,bauza2013traffic,sundar2014implementing}, especially on cooperative vehicular systems, such as vehicle-to-vehicle communications. Congestion patterns are introduced to quantify the congestion, either defined or learned in terms of discretized representations~\cite{horiguchi2004effective,zhang2011vehicle,xu2013identifying,zhang2017analyzing,xiong2018predicting}. In this paper, we exploit the congestion patterns as contextual cues for multi-agent trajectory prediction.

\paragraph*{\ac{gnn} for Trajectory Prediction} 

With a powerful ability in representation learning, \acs{gnn} \cite{scarselli2008graph,li2015gated,velivckovic2017graph} saliently improve the performance on diverse tasks. Recently, researchers start to adopt \ac{gnn} for trajectory prediction; both \ac{gat}~\cite{kosaraju2019social,huang2019stgat} and \ac{gcn}~\cite{sun2020recursive} have been used to aggregate information from all or neighboring agents. These works treat each agent as a graph node with various ways to construct weights of graph edges~\cite{vemula2018social,eiffert2019predicting,choi2019drogon,ivanovic2019trajectron}. In comparison, the proposed method constructs a graph where each edge explicitly encodes an \ac{sf}-based distance among all agents and recruits a \ac{gcn} to learn congestion.

\setstretch{1}

\section{Methods}\label{sec:method}

This section describes the proposed method of congestion-aware multi-agent trajectory prediction; see \cref{fig:architecture} for an overview. We start with a brief problem definition of multi-agent trajectory prediction in \cref{sec:problem}, following the conventions in~\cite{alahi2016social}. Next, we introduce the learning of congestion patterns in multi-agent scenarios and describe how a teacher learns the contextual cues in \cref{sec:congestion_pattern}. In \cref{sec:pattern_matching}, we formulate the optimization problem of congestion pattern matching so that a student can reason about congestion patterns taught by the teacher. We jointly solve this optimization problem with trajectory prediction by proposing a generic encoder-pooling-decoder model that predicts future collision-free trajectories in \cref{sec:trajectory_prediction}.

\subsection{Problem Definition}\label{sec:problem}

The goal of multi-agent trajectory prediction is to predict the future trajectories of all on-scene agents given their trajectory histories. We denote $\{\zeta^m, m=1,...,n\}$ as the set of trajectories for all $n$ agents. At time step $t$, the position of the $m$th agent is represented by its local 2D coordinates $\zeta^m_t=(x^m_t, y^m_t)$. Given a time span $t=1:T_h$, the observation is denoted as history trajectories $\zeta_h = \{\zeta^m_{t=1:T_h}\}$, and the future trajectories up to the time step $T_p$ is denoted as $\zeta_p = \{\zeta^m_{t=T_h+1:T_p}\}$. In short, the multi-agent trajectory prediction is formulated as  estimating the probability $P(\zeta_p | \zeta_h)$.

\subsection{Learning Congestion Patterns}\label{sec:congestion_pattern}

Intuitively, congestion patterns among agents in multi-agent navigation scenarios provide crucial contextual cues for trajectory prediction; they not only describe the position and intention of the agents, but also present safety-critical information about collisions. In literature, congestion patterns~\cite{xu2013identifying,xiong2018predicting} are mostly described by empirical equations about the relations of vehicle positions or clustering algorithms that group the observations into pre-defined categories. Such definitions have obvious shortcomings; for instance, it is sensitive with respect to the number of agents, vehicle moving velocities, \etc. In fact, it is non-trivial to quantify the congestion patterns by an explicit set of rules or formulae. Instead, given the trajectory history as observation $o = \zeta_h$, we propose to learn the congestion patterns unsupervisedly: We use graph-based generative learning to derive the hidden congestion patterns and build a probabilistic \ac{gmm} to account for various modes.

\paragraph*{Graph Representation}

To capture the congestion patterns embedded in physics constraints, we build the graph in the following way. Given two agents $u, v \in \mathcal{V}_t$, the graph $A_t = (\mathcal{V}_t, \mathcal{E}_t)$ at each frame $t \in \{1,..,T_h\}$ is constructed by their 2D locations $(x_t, y_t)$ and velocities $(\dot{x}_t, \dot{y}_t)$. The graph adjacency matrix $\mathcal{E}_t = \{\mathcal{E}^{uv}_t\}, u, v = 1,..,n$ is defined as:
\begin{equation}
\small
    \mathcal{E}^{uv}_t = \mathcal{E}^{vu}_t = 
    \begin{cases}
        1 / t^{uv}_c, t^{uv}_c > 0 \\
        0, t^{uv}_c = 0
    \end{cases},
    \label{equ:graph}
\end{equation}
where the estimated collision time is $t^{uv}_c = \max(-\frac{\Delta^{uv} x \times \Delta^{uv} \dot{x} + \Delta^{uv} y \times \Delta^{uv} \dot{y}}{{\Delta^{uv} \dot{x}}^2 + {\Delta^{uv} \dot{y}}^2}, 0)$, and $\Delta^{uv}(\cdot)$ denotes the quantity difference between agents $u$ and $v$. Intuitively, a larger weight reflects a higher chance of collision, and the matrix describes the scene and congestion conditions.

\paragraph*{Generative Learning}

We leverage \ac{vae}~\cite{kingma2013auto} to unsupervisedly learn the latent congestion pattern $z$ in the graphs. Specifically, we follow the graph \ac{vae} approach proposed in \ac{gcn}~\cite{kipf2016semi} where both the encoder and decoder are instantiated as graph convolutional layers. The objective is to optimize the reconstructed graph representation $A_t$ while regularizing the latent distribution.

\paragraph*{\acf{gmm}}

As the congestion patterns are naturally multi-modal, we further use a Gaussian Mixture Model to account for various modes. Specifically, treating the latent congestion pattern $z$ as a random variable~\cite{dilokthanakul2016deep,jiang2016variational,yang2019deep}, we build the \ac{gmm} as:
\begin{equation}
\small
    \mathcal{Q}(z) = \sum_i^{M_\mathcal{Q}} \lambda_i q_i(z),
    \label{equ:Q_dist}
\end{equation}
where each mixture component $q_i(z)$ is a Gaussian distribution, $\lambda_i$ is the mixture weight, and $M_\mathcal{Q}$ is the hyperparameter specifying the total number of mixtures. The mixture model can be learned by the stochastic EM~\cite{celeux1985sem} algorithm. As the hidden congestion pattern $z$ is extracted from the observation $o$, we denote the mixture model as $\mathcal{Q}(o)$ henceforth.

\subsection{Matching Congestion Patterns}\label{sec:pattern_matching}

While generating the trajectories, we hope that the student model can simultaneously match the congestion patterns taught by the teacher, such that the predicted trajectories are collision-free. We will first describe how to match the student's congestion model and that of the teacher's and defer the implementation details to the next section. Denoting the student's congestion pattern model as $\mathcal{P}(o)$, congestion pattern matching can be formulated as the $\kl$-divergence between two pattern distributions:
\begin{equation}
\small
    \min \mathbb{D}_{\kl}(\mathcal{P}(o) \Vert \mathcal{Q}(o)).
    \label{equ:KL}
\end{equation}
Considering the mixture nature of $\mathcal{Q}(o)$, we model $\mathcal{P}(o)$ also as a Gaussian mixture, \ie, $\mathcal{P}(o) = \sum_j^{M_\mathcal{P}} \omega_j p_j(o)$, where the total number of mixture $M_\mathcal{P}$ could be different from $M_\mathcal{Q}$. 

Since there is no analytical solution for \cref{equ:KL}, we solve it by optimizing a variational upper bound. Similar to~\cite{hershey2007approximating,xie2019representation}, we propose a variational parametrization approach to solve the optimization problem. By decomposing the mixture weights $\omega_j = \sum_i^{M_\mathcal{Q}}\alpha_{ij}$ and $\lambda_i = \sum_j^{M_\mathcal{P}}\beta_{ij}$, the objective in \cref{equ:KL} can be rewritten as:
\begin{equation}
\fontsize{8}{8}\selectfont
\begin{aligned}
    \mathbb{D}_{\kl}(\mathcal{P}(o) \Vert \mathcal{Q}(o))
    &= -\int \mathcal{P}(o)\log\frac{1}{\mathcal{P}(o)}\left(\sum_{i, j}\beta_{ij}q_i(o)\right) \\
    &= -\int \mathcal{P}(o)\log\frac{1}{\mathcal{P}(o)}\left(\sum_{i, j}\frac{\beta_{ij}q_i(o) \alpha_{ij}p_j(o)}{\alpha_{ij}p_j(o)}\right).
    \label{equ:KL_expand}
\end{aligned}
\end{equation}
Using Jensen's inequality, \cref{equ:KL_expand} can be transformed to:
\begin{equation}
\small
\begin{aligned}
    \mathbb{D}_{\kl}(\mathcal{P}(o) \Vert \mathcal{Q}(o))
    &\le -\int \mathcal{P}(o)\sum_{i, j}\frac{\alpha_{ij}p_j(o)}{\mathcal{P}(o)} \log\frac{\beta_{ij}q_i(o)}{\alpha_{ij}p_j(o)} \\
    &= \sum_{i, j}\alpha_{ij}\mathbb{D}_{\kl}(p_j(o) \Vert q_i(o))+D_{\kl}(\alpha \Vert \beta).
    \label{equ:KL_inequality}
\end{aligned}
\end{equation}

\setstretch{0.99}

We optimize \cref{equ:KL} by minimizing its upper bound:
\begin{equation}
\small
    \min_{\{p_j\},\alpha,\beta} L_1 = \sum_{i, j}\alpha_{ij}\mathbb{D}_{\kl}(p_j(o) \Vert q_i(o)) + \mathbb{D}_{\kl}(\alpha \Vert \beta).
    \label{equ:KL_new}
\end{equation}
Note that the convergence of the optimization problem has been guaranteed as discussed in~\cite{hershey2007approximating}.

To solve \cref{equ:KL_new}, we iteratively optimize $\{p_j\}$, $\alpha$, and $\beta$. Assuming fixed $\alpha$ and $\beta$,
\begin{equation}
\small
\begin{aligned}
    &\min_{\{p_j\}}\sum_{i, j}\alpha_{ij}\mathbb{D}_{\kl}(p_j(o) \Vert q_i(o))\\ 
    &= \sum_{i, j}\alpha_{ij}\left(\mathbb{E}_{p_j(o)}[-\log q_i(o)] - \mathbb{H}[p_j(o)]\right).
    \label{equ:KL_first}
\end{aligned}
\end{equation}
With $\{p_j\}$ learned, $\alpha$ and $\beta$ can be updated by the closed-form solutions:
\begin{equation}
\small
    \alpha_{ij} = \frac{\omega_j \beta_{ij} \exp^{-\mathbb{D}_{\kl}(p_j(o) \Vert q_i(o))}}{\sum_{i^\prime} \beta_{i^\prime j} \exp^{-\mathbb{D}_{\kl}(p_j(o) \Vert q_{i^\prime}(o))}}, \quad
    \beta_{ij} = \frac{\lambda_i \alpha_{ij}}{\sum_{j^\prime} \alpha_{ij^\prime}}.
    \label{equ:KL_second}
\end{equation}
The overall algorithm for the above congestion pattern matching (CPM) process is summarized in \cref{alg:CPM}. Please refer to supplementary video for a full derivation.

\begin{algorithm}[ht!]
\small
    \caption{Congestion Pattern Matching (CPM)}
    \label{alg:CPM}
    \begin{algorithmic}[1]
        \STATE Input: the learned congestion patterns $\mathcal{Q}(o)$
        \STATE Initialize $\alpha_{ij}$ and $\beta_{ij}$
        \WHILE{not converged}
            \STATE Fix $\alpha_{ij}$ and $\beta_{ij}$ and optimize $\{p_j\}$ using \cref{equ:KL_first}
            \STATE Fix $\{p_j\}$ and update $\alpha_{ij}$ and $\beta_{ij}$ using \cref{equ:KL_second}
        \ENDWHILE
    \end{algorithmic}
\end{algorithm}

\subsection{Collision-free Trajectory Prediction}\label{sec:trajectory_prediction}

We make the student jointly predict trajectories and match the teacher's congestion patterns. As illustrated in \cref{fig:architecture}, the trajectory prediction in the student model comprises an encoder module that encodes observed trajectories, a pooling module that models the spatial relations among agents, and a decoder module that recursively generates the future trajectories. The output of the social features at the pooling module is taken to match (distribution matching; see \cref{equ:KL_new}) the teacher model's congestion pattern.

Our proposed student model can be trained end-to-end by iteratively minimizing the congestion pattern matching loss in \cref{equ:KL_new} and the trajectory prediction loss, defined as
\begin{equation}
    L_2 = -\frac{1}{m} \sum_m \sum_{t=T_h+1:T_p} \log P(\zeta^m_{p_t}|\zeta^m_h),
    \label{equ:traj}
\end{equation}
where $\zeta^m_h$ and $\zeta^m_p$ are the observed and predicted trajectories.

\paragraph*{Implementation Details}

The teacher model is composed of the GCN-VAE architecture with a latent dimension of $64$. The \ac{gmm} is a deep learned from the latent. We use fully connected layers to represent the congestion pattern. The teacher model is trained using Adam~\cite{kingma2014adam} with a learning rate of $1\times10^{-4}$. For the student model, it is compatible with prevalent encoder-pooling-decoder architectures~\cite{deo2018convolutional,alahi2016social}. The pooling module is implemented following the social convolution pooling~\cite{deo2018convolutional}. The encoder and decoder modules are created using LSTMs with a fixed hidden dimension of size $128$. The $\rm{CPM}(\cdot)$ module is implemented as another deep \ac{gmm}, which outputs the parameters of each mixture component. The number of components is a tunable hyper-parameter; see \cref{sec:qualitative}. The student model is learned using Adam~\cite{kingma2014adam} with a learning rate of $3\times10^{-3}$. Both models are implemented in PyTorch~\cite{paszke2019pytorch}.

\section{Experiments}\label{sec:evaluation}

\subsection{Datasets}\label{sec:datasets}

\paragraph*{GTA Dataset}

\begin{figure}[hbt!]
    \centering
    \begin{subfigure}[c]{0.5\linewidth}
        \includegraphics[width=\linewidth]{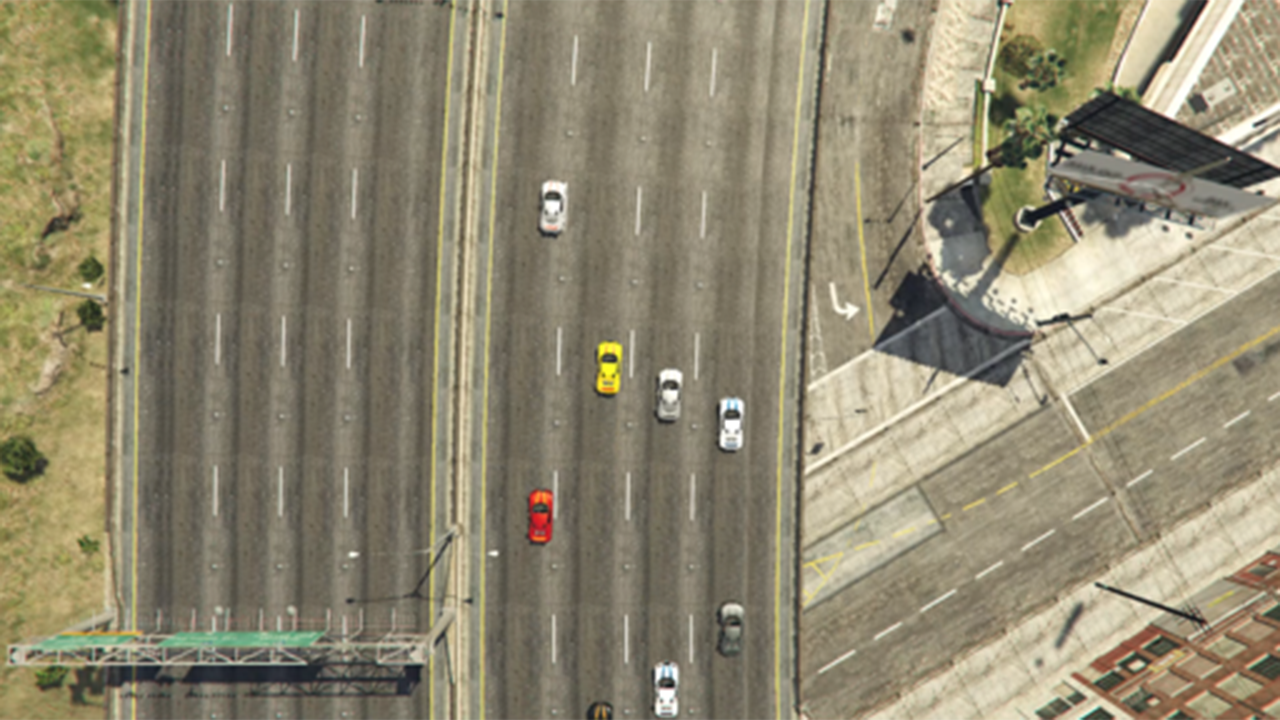}
        \caption{Scenario 1}
    \end{subfigure}%
    \begin{subfigure}[c]{0.5\linewidth}
        \includegraphics[width=\linewidth]{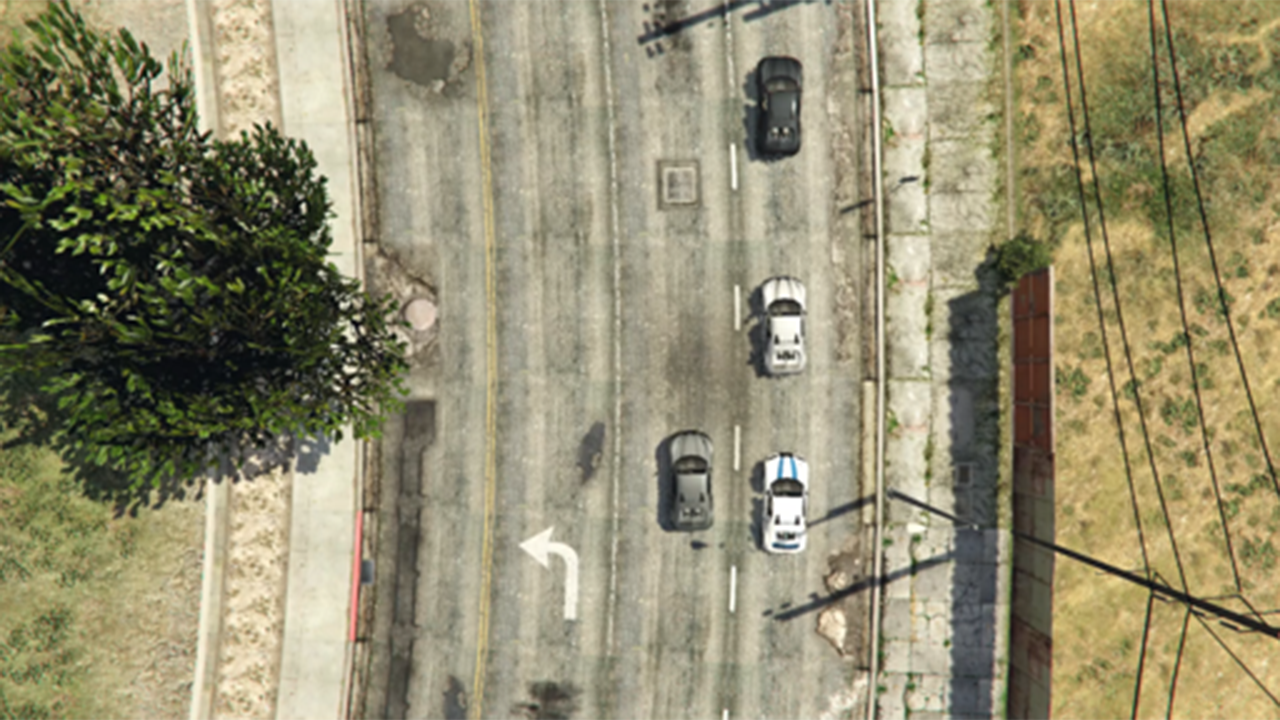}
        \caption{Scenario 2}
    \end{subfigure}%
    \\
    \begin{subfigure}[c]{0.5\linewidth}
        \includegraphics[width=\linewidth]{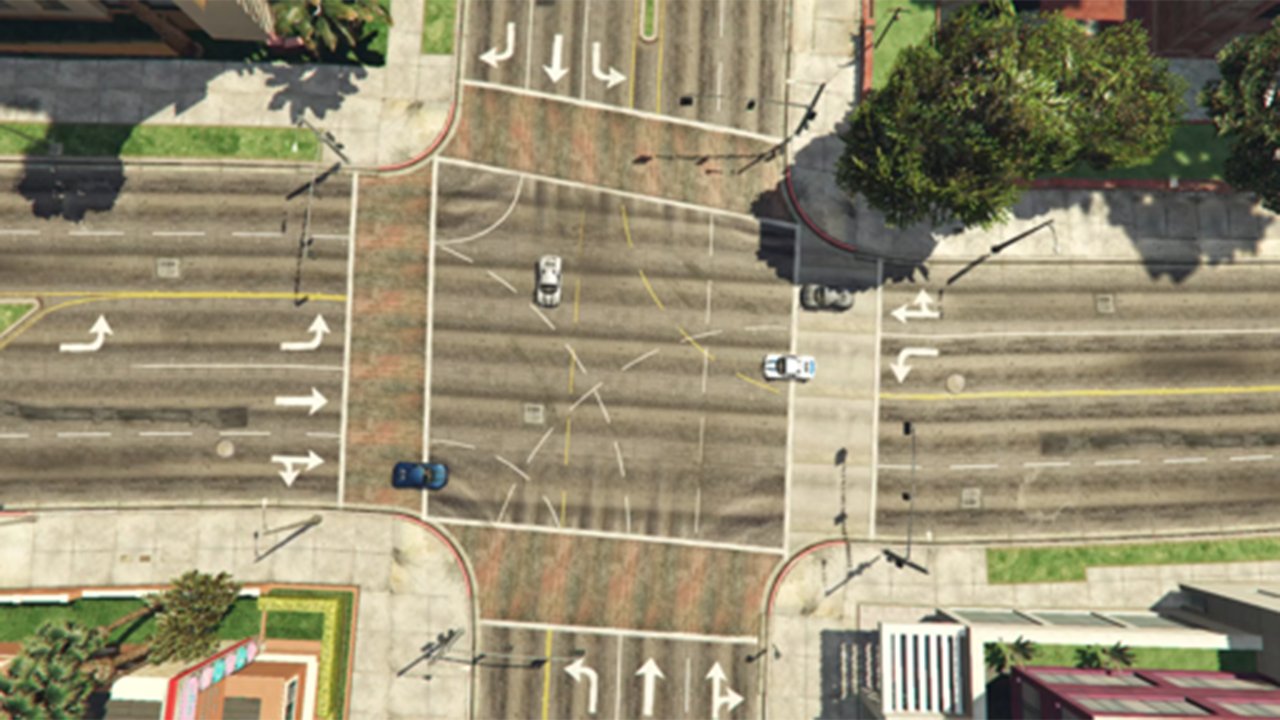}
        \caption{Scenario 3}
    \end{subfigure}%
    \begin{subfigure}[c]{0.5\linewidth}
        \includegraphics[width=\linewidth]{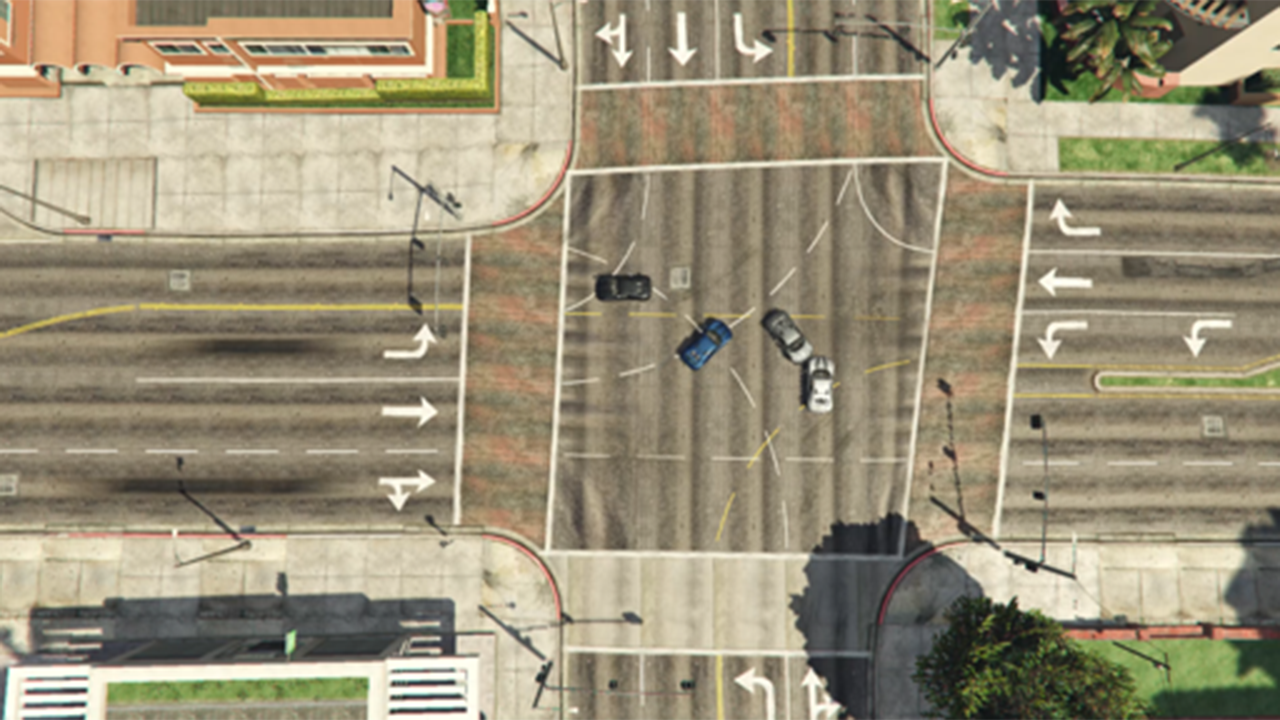}
        \caption{Scenario 4}
    \end{subfigure}%
    \caption{Sample top-views of the four scenarios in our GTA dataset.}
    \label{fig:dataset}
\end{figure}

To evaluate collision avoidance, we create a novel dataset based on the popular game platform of Grand Theft Auto (GTA). Compared to other platforms~\cite{xie2019vrgym,airsim2017fsr,dosovitskiy17} that supports multi-agent simulations, the GTA models realistic urban-scale traffic commuting system; see \cref{table:gta_dataset} for other dataset statistics. This dataset focuses on trajectory prediction under safety-critical scenarios with rich vehicle interactions. By developing modding scripts, we create four types of safety-critical scenarios (see \cref{fig:dataset}): (i) highway vehicle driving (mainly vehicle following), (ii) local vehicle driving (overtaking can frequently happen), (iii) driving in intersections (no traffic rules and crowded driving scenarios), and (iv) aggressive behaviors (almost lead to collisions). We use the four types of driving scenarios to study collision-free trajectory prediction. In experiments, we split the entire dataset into 3 folds for training and 1 fold for testing. All trajectories contain 3s of observations and 5s of predictions, and a model is tasked to predict agents' future paths.

\begin{table}[ht!]
    \caption{GTA dataset statistics.}
    \begin{center}
        \resizebox{\linewidth}{!}{
        \begin{tabular}{c c c c}
            \toprule
            Total Clips & Vehicle Trajectories & Highway Trajectories & Local Trajectories \\
            \midrule
            3300 & 27813 & 18229 & 9584 \\
            \toprule
            Following Events & Overtaking Events & Collision Events & \\
            \midrule
            7055 & 2300 & 890 & \\
            \bottomrule
        \end{tabular}
        }
    \end{center}
    \label{table:gta_dataset}
\end{table}

\paragraph*{NGSIM Dataset}

We also evaluate the accuracy of trajectory prediction on the commonly used NGSIM US-101~\cite{colyar2007us} dataset to show the competitiveness of the proposed method. The dataset contains real highway traffic data that is captured over a time span of 45 minutes. Similar to the GTA dataset, We split the trajectories into 8s segments where 3s are used for observations and 5s for predictions.

\subsection{Baselines and Evaluation Metrics}\label{sec:metrics}

We compare our model with several well-established baselines~\cite{deo2018convolutional,tang2019multiple,zhao2019multi,gupta2018social,alahi2016social,schmerling2018multimodal,deo2018would,ho2016generative,bhattacharyya2018multi,bhattacharyya2019simulating} and report the following metrics results:

\textbf{Collision Rate} evaluates the performance of collision avoidance for the predicted trajectories. In the GTA dataset, we calculate the collision rate by counting the collision events among trajectories and divide it by the total number of trajectories for all trials. The ground-truth collision event is obtained from the game simulator.

\textbf{Root Mean Squared Error (RMSE)} evaluates the accuracy of the predicted trajectories, calculated across trajectories given different time horizons (1s-5s). Note that for baseline models that generate trajectories using GANs~\cite{zhao2019multi,gupta2018social}, we sample $k$ output predictions for each trial and choose the ``best'' prediction in the sense of $L_2$ norm for evaluation.

\subsection{Quantitative Results}\label{sec:quantitative}

\paragraph*{Results on GTA Dataset}

As shown in \cref{table:gta_colli_results,table:gta_results}, the proposed \textbf{CF-LSTM} achieves the best performance on the collision rate and RMSE in the GTA dataset, proving its strength in collision-free trajectory prediction. Specifically, for the collision rate, CF-LSTM demonstrates the lowest error rates across all scenarios. Of the four scenarios, we note that Scenario 4 is the most challenging in the sense that there is no assumption of rational driving. Moreover, we notice a higher chance of collision in crowded space, \eg, intersections where vehicles meet with each other. For RMSE, CF-LSTM attains the best result as presented in \cref{table:gta_results}, with the minimum average RMSE compared to other baseline methods. As shown in the table, the RMSE metric alone does not tell the difference between Scenario 3 and Scenario 4 for all the methods, which indicates that the accuracy of trajectory prediction does not perfectly reflect driving safety. We argue that there should be more effective metrics, such as the collision rate, to measure it.

\begin{table}[ht!]
    \caption{Collision rate (\%) on GTA}
    \label{table:gta_colli_results}
    \centering
    \resizebox{\linewidth}{!}{
        \begin{tabular}{l c c c c c c}
            \toprule
            Methods & V-LSTM & CS-LSTM~\cite{deo2018convolutional} & S-GAN~\cite{gupta2018social} & CF-LSTM (Ours) \\
            \midrule
            Scenario 1 & 4.219 & 3.086 & 3.372 & \textbf{2.909} \\
            Scenario 2 & 5.830 & 4.345 & \textbf{4.015} & 4.170 \\
            Scenario 3 & 8.331 & 6.997 & 5.805 & \textbf{5.397} \\
            Scenario 4 & 11.676 & 9.500 & 8.923 & \textbf{8.766} \\
            Avg & 7.514 & 5.982 & 5.529 & \textbf{5.310} \\
            \bottomrule
        \end{tabular}
    }%
\end{table}
\begin{table}[ht!]
    \caption{RMSE on GTA.}
    \label{table:gta_results}
    \centering
    \resizebox{\linewidth}{!}{
        \begin{tabular}{l c c c c c}
            \toprule
            Methods & V-LSTM & CS-LSTM~\cite{deo2018convolutional} & S-GAN~\cite{gupta2018social} & CF-LSTM (Ours) \\
            \midrule
            Scenario 1 & 1.88 & 1.25 & 1.40 & \textbf{1.11} \\
            Scenario 2 & 1.91 & 1.84 & \textbf{1.74} & 1.76 \\
            Scenario 3 & 2.98 & 2.55 & 2.67 & \textbf{2.42} \\
            Scenario 4 & 3.02 & 2.89 & 2.96 & \textbf{2.76} \\
            Avg & 2.45 & 2.13 & 2.19 & \textbf{2.01} \\
            \bottomrule
        \end{tabular}
    }
\end{table}

\paragraph*{Results on NGSIM Dataset}

\cref{table:ngsim_results} shows performance of various models on the NGSIM dataset. The proposed \textbf{CF-LSTM} significantly outperforms the deterministic physics-based models of CV and C-VGMM+VIM~\cite{deo2018would} and surpasses planning-based models, such as GAIL-GRU~\cite{ho2016generative} and PS-GAIL~\cite{bhattacharyya2018multi,bhattacharyya2019simulating}, and pattern-based models, such as V-LSTM, S-LSTM~\cite{alahi2016social}, and CS-LSTM~\cite{deo2018convolutional}. CF-LSTM improves the previous state-of-the-art in three settings compared to MFP~\cite{tang2019multiple}. However, the latter needs additional scene semantics for prediction in every time step while CF-LSTM does not. Our method also fares better than MATF GAN~\cite{zhao2019multi}. These results verify the competitiveness of CF-LSTM.

\begin{table*}[hbt!]
    \caption{RMSE on NGSIM.}
    \begin{center}
        \resizebox{\linewidth}{!}{
        \begin{tabular}{l c c c c c c c c c c c}
            \toprule
            Times(s) & CV & C-VGMM+VIM~\cite{deo2018would} & V-LSTM & S-LSTM~\cite{alahi2016social} & CS-LSTM~\cite{deo2018convolutional} & MFP~\cite{tang2019multiple} & MATF GAN~\cite{zhao2019multi} & VAE & GAIL-GRU~\cite{ho2016generative} & PS-GAIL~\cite{bhattacharyya2018multi,bhattacharyya2019simulating} & CF-LSTM (Ours) \\
            \midrule
            1s & 0.73 & 0.66 & 0.66 & 0.65 & 0.61 & \textbf{0.54} & 0.66 & 0.68 & 0.69 & 0.60 & 0.55 \\
            2s & 1.78 & 1.56 & 1.64 & 1.31 & 1.27 & 1.16 & 1.34 & 1.72 & 1.51 & 1.83 & \textbf{1.10} \\
            3s & 3.13 & 2.75 & 2.94 & 2.16 & 2.09 & 1.89 & 2.08 & 2.77 & 2.55 & 3.14 & \textbf{1.78} \\
            4s & 4.78 & 4.24 & 4.59 & 3.25 & 3.10 & 2.75 & 2.97 & 3.94 & 3.65 & 4.56 & \textbf{2.73} \\
            5s & 6.68 & 5.99 & 6.60 & 4.55 & 4.37 & \textbf{3.78} & 4.13 & 5.21 & 4.71 & 6.48 & 3.82 \\
            \bottomrule
        \end{tabular}
        }
    \end{center}
    \label{table:ngsim_results}
\end{table*}

\setstretch{0.94}

\subsection{Qualitative Results}\label{sec:qualitative}

\begin{figure*}[t!]
    \centering
    \begin{subfigure}[c]{0.248\linewidth}
        \begin{overpic}[width=\linewidth]{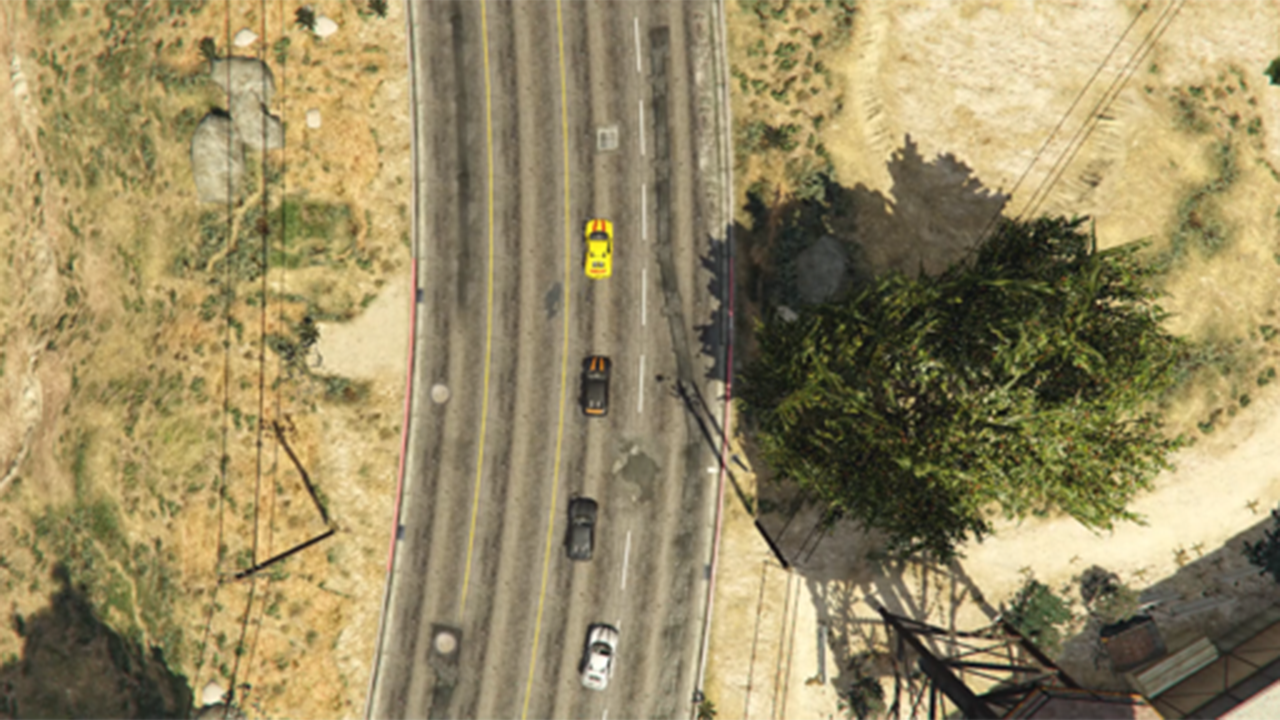}
            \put(57,1){\color{white}\linethickness{0.5mm}%
                \frame{\includegraphics[width=0.42\linewidth]{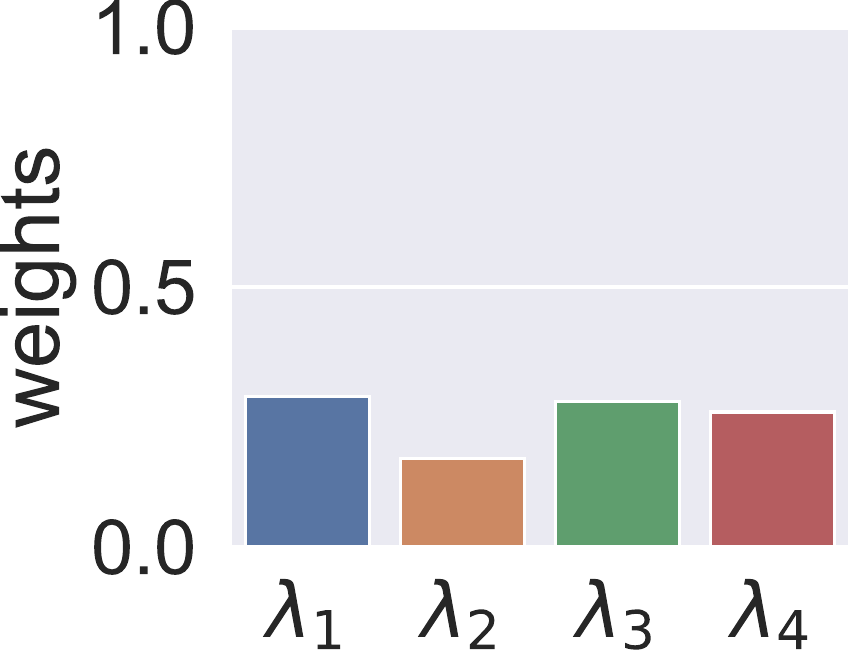}}
            }
        \end{overpic}
        % \caption{Frame \#1}
    \end{subfigure}%
    \hfill
    \begin{subfigure}[c]{0.248\linewidth}
        \begin{overpic}[width=\linewidth]{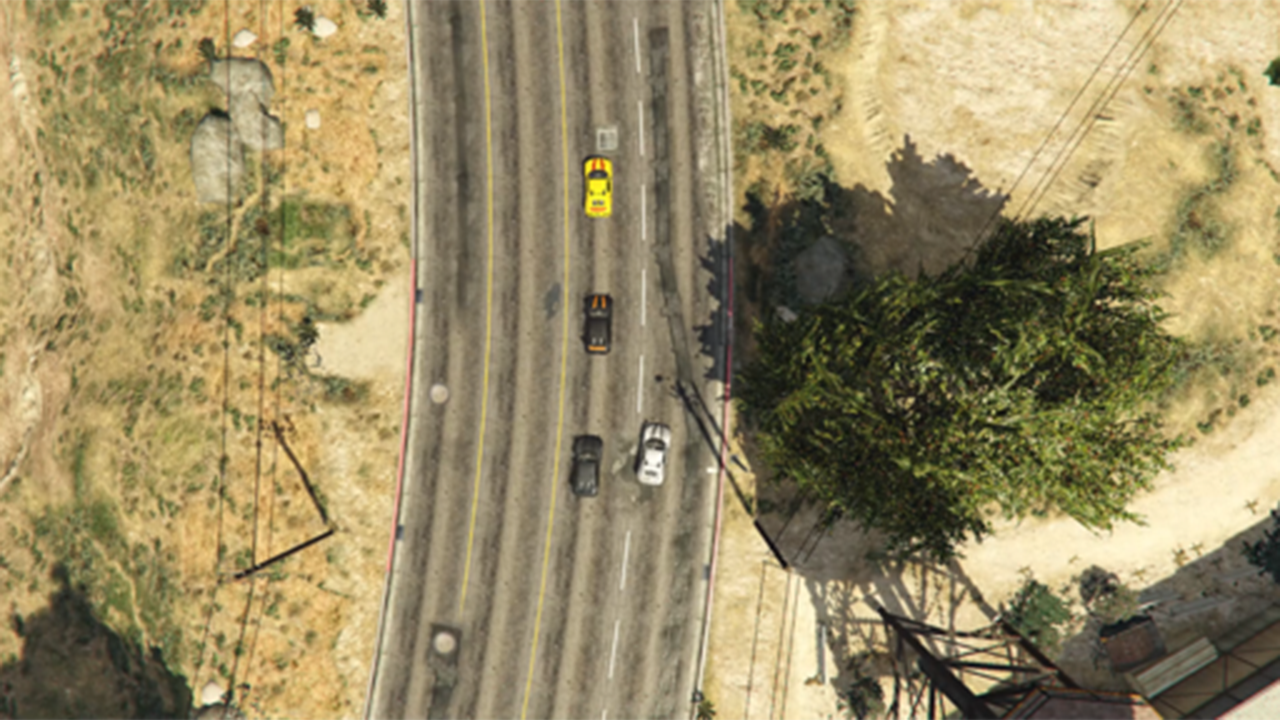}
            \put(57,1){\color{white}\linethickness{0.5mm}%
                \frame{\includegraphics[width=0.42\linewidth]{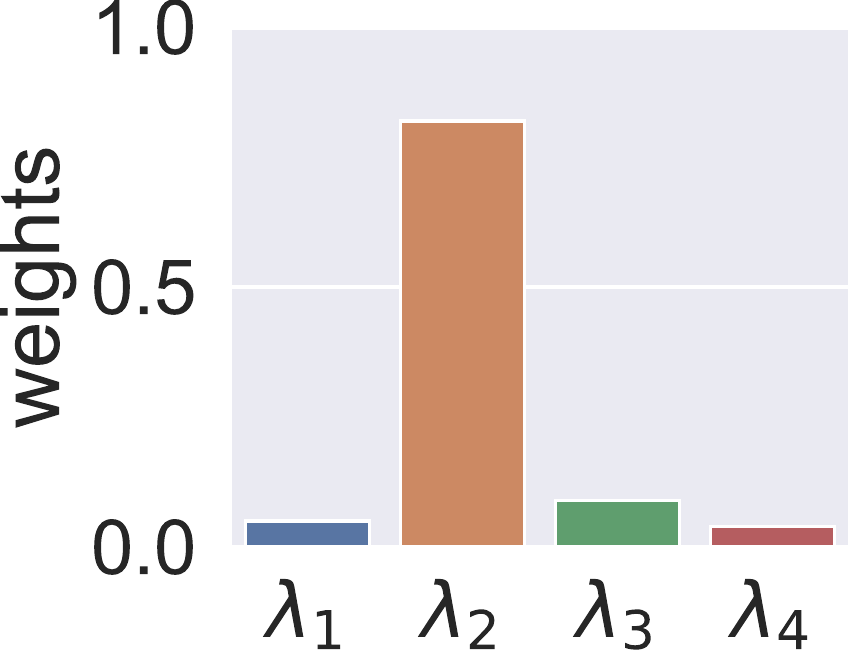}}
            }
        \end{overpic}
        % \caption{Frame \#2}
    \end{subfigure}%
    \hfill
    \begin{subfigure}[c]{0.248\linewidth}
        \begin{overpic}[width=\linewidth]{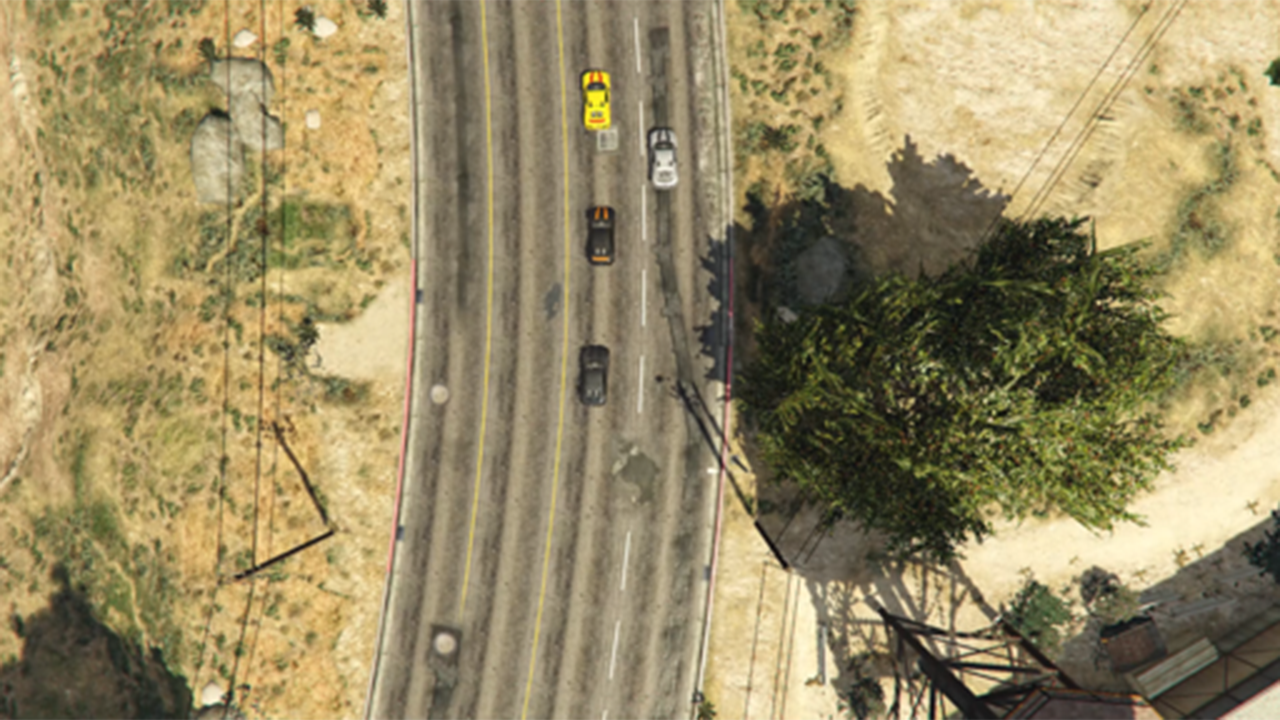}
            \put(57,1){\color{white}\linethickness{0.5mm}%
                \frame{\includegraphics[width=0.42\linewidth]{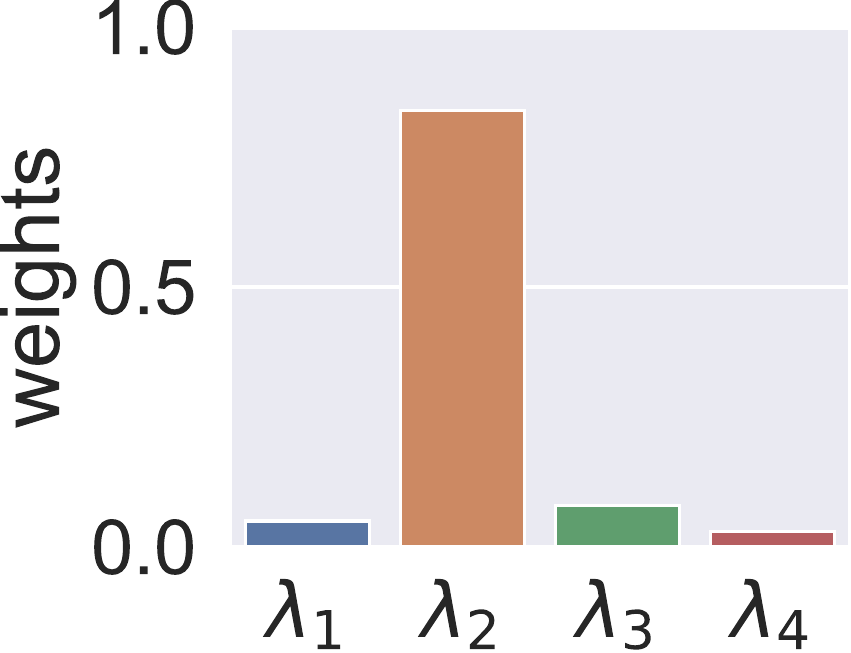}}
            }
        \end{overpic}
        % \caption{Frame \#3}
    \end{subfigure}%
    \hfill
    \begin{subfigure}[c]{0.248\linewidth}
        \begin{overpic}[width=\linewidth]{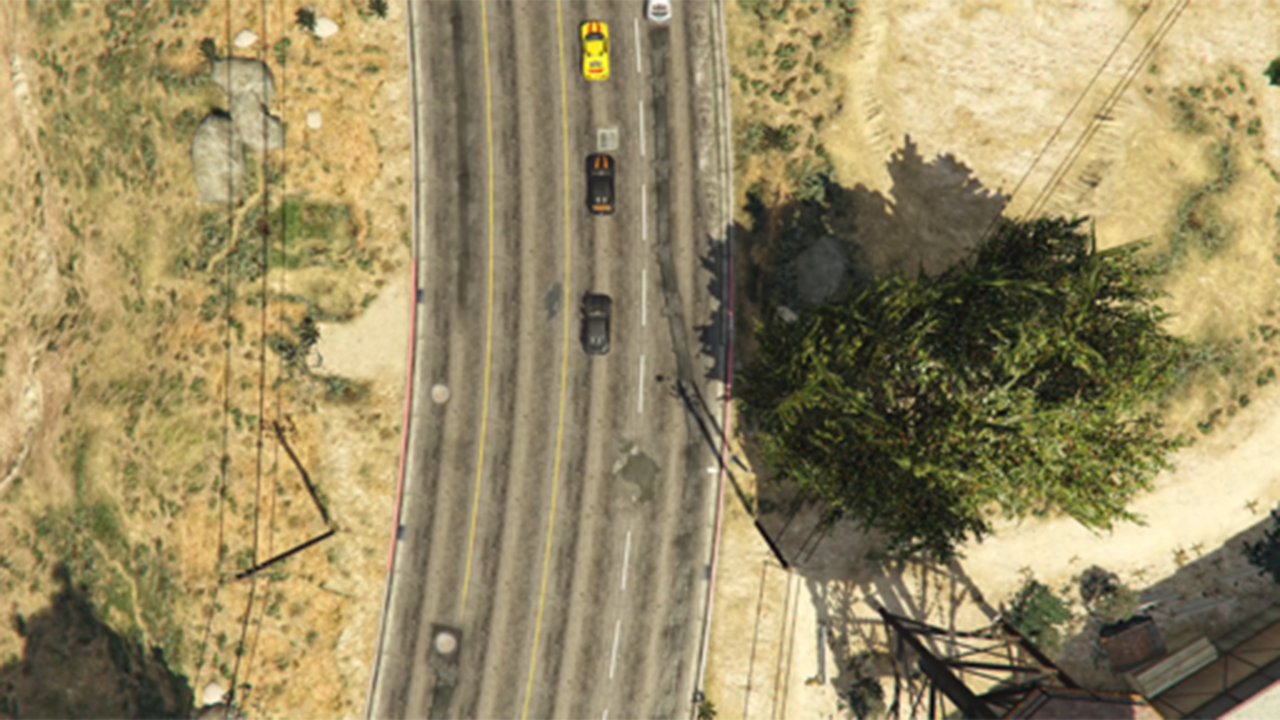}
            \put(57,1){\color{white}\linethickness{0.5mm}%
                \frame{\includegraphics[width=0.42\linewidth]{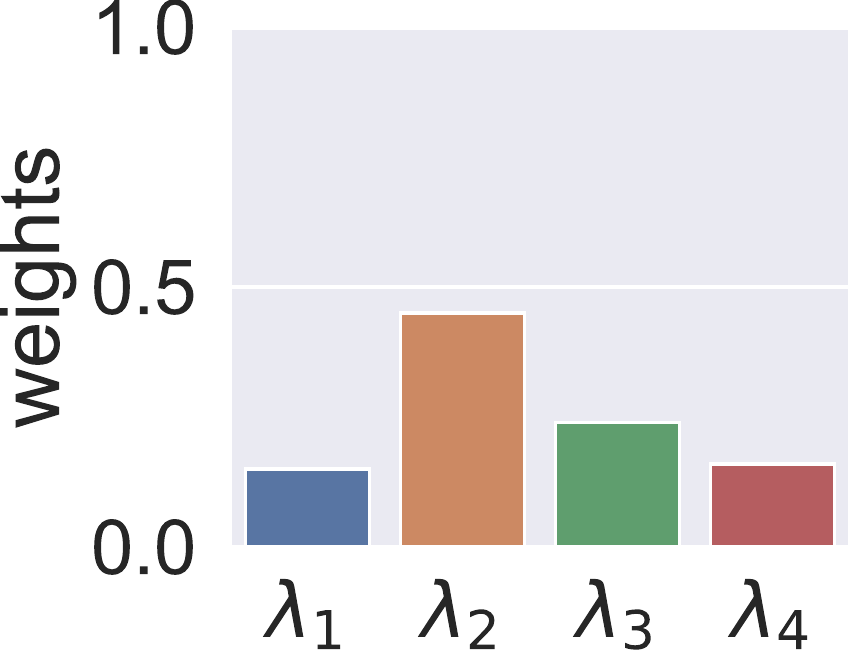}}
            }
        \end{overpic}
        % \caption{Frame \#4}
    \end{subfigure}%
    \\\vspace{1pt}
    \begin{subfigure}[c]{0.248\linewidth}
        \begin{overpic}[width=\linewidth]{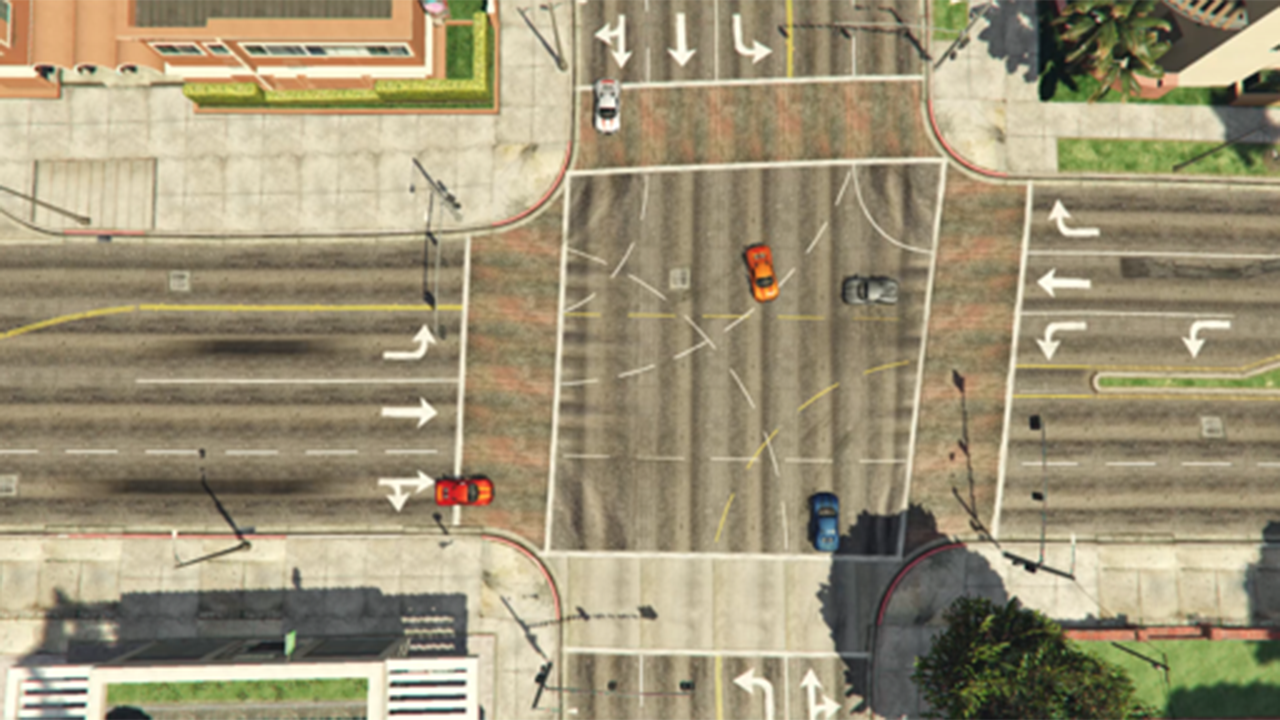}
            \put(1,23){\color{white}\linethickness{0.5mm}%
                \frame{\includegraphics[width=0.42\linewidth]{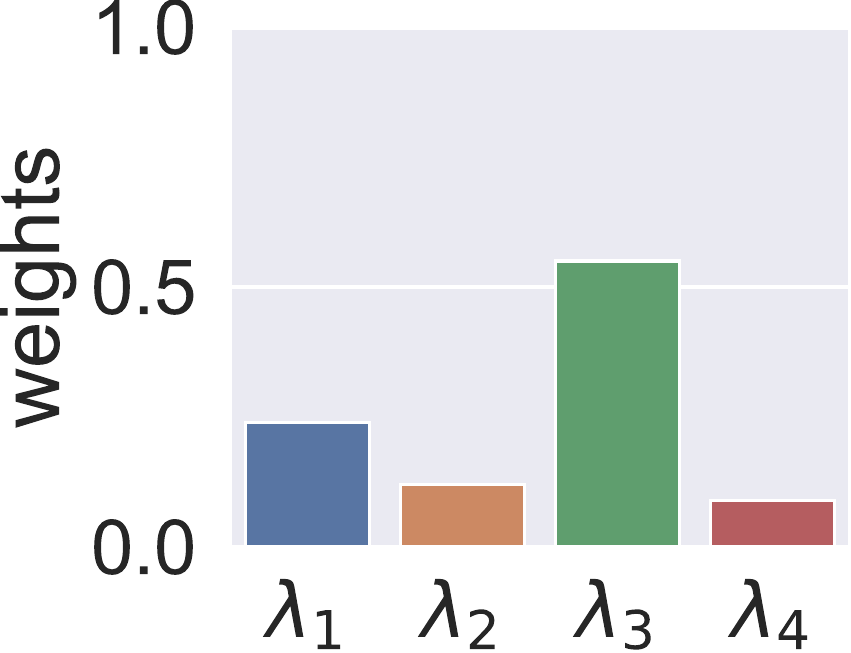}}
            }
        \end{overpic}
    \end{subfigure}%
    \hfill
    \begin{subfigure}[c]{0.248\linewidth}
        \begin{overpic}[width=\linewidth]{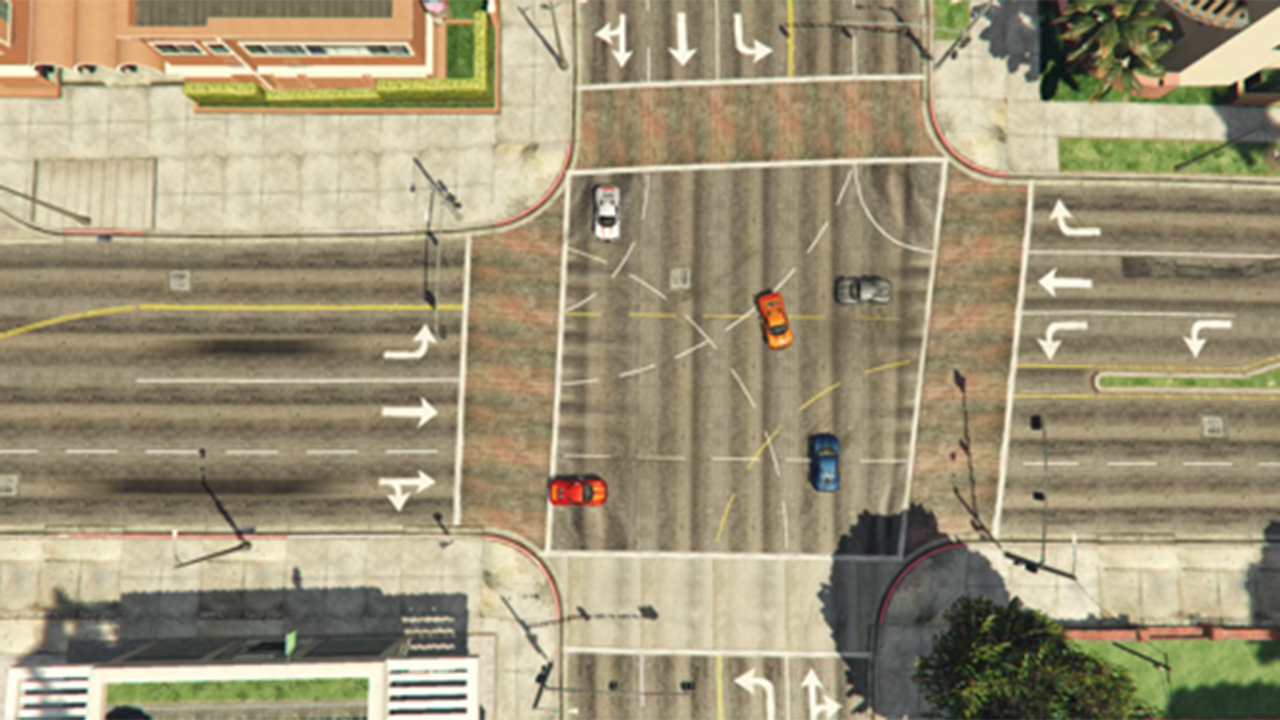}
            \put(1,23){\color{white}\linethickness{0.5mm}%
                \frame{\includegraphics[width=0.42\linewidth]{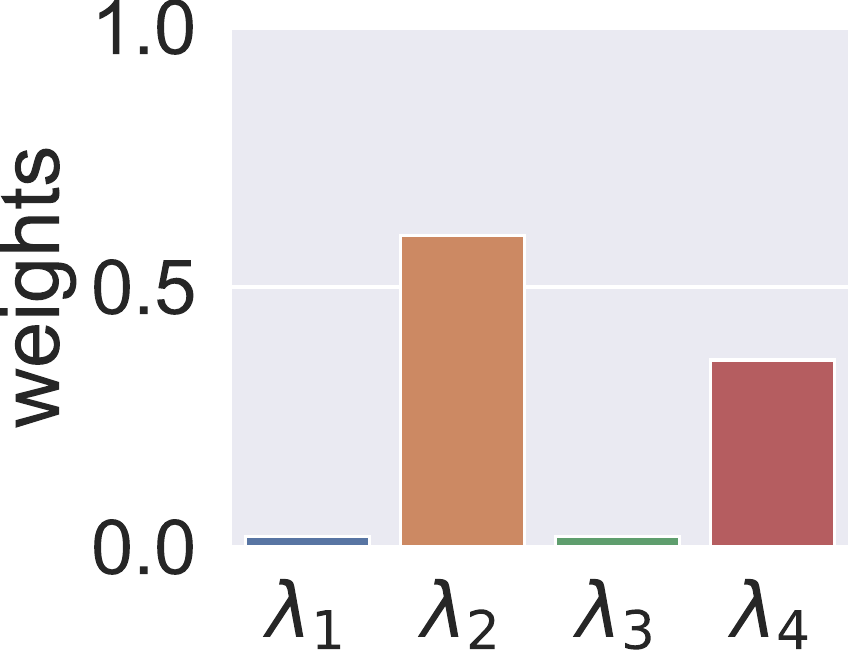}}
            }
        \end{overpic}
    \end{subfigure}%
    \hfill
    \begin{subfigure}[c]{0.248\linewidth}
        \begin{overpic}[width=\linewidth]{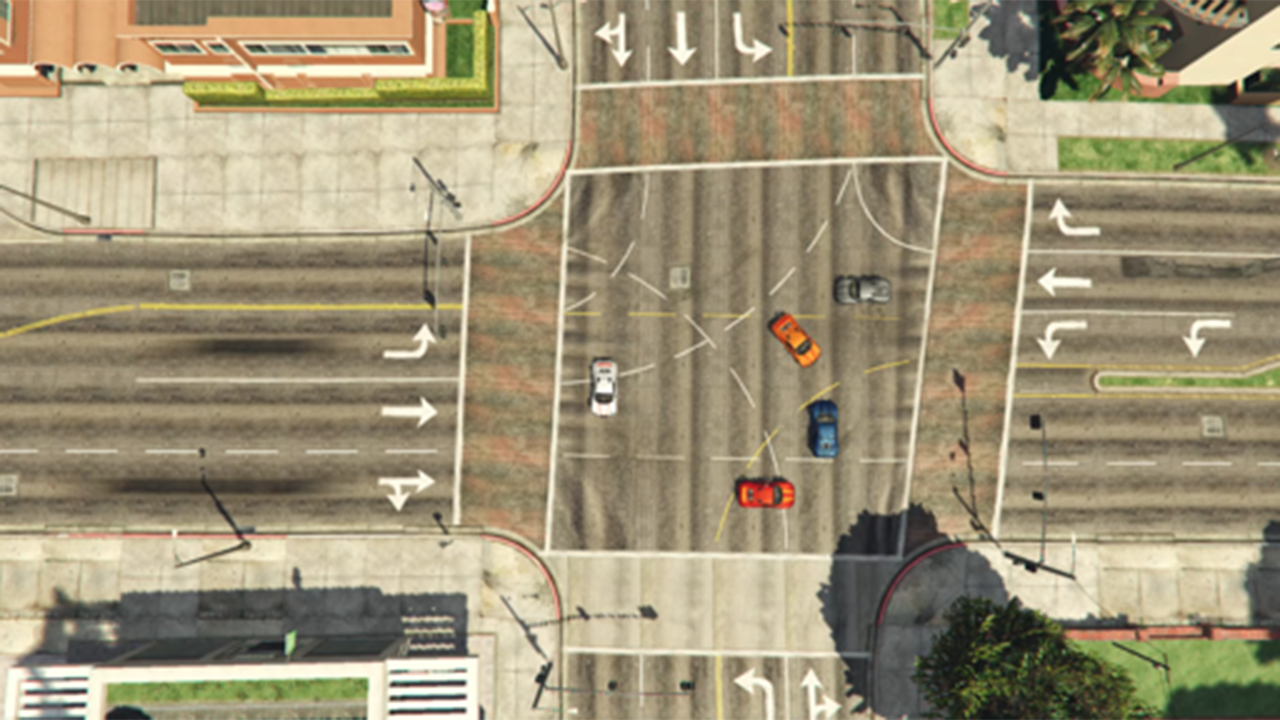}
            \put(1,23){\color{white}\linethickness{0.5mm}%
                \frame{\includegraphics[width=0.42\linewidth]{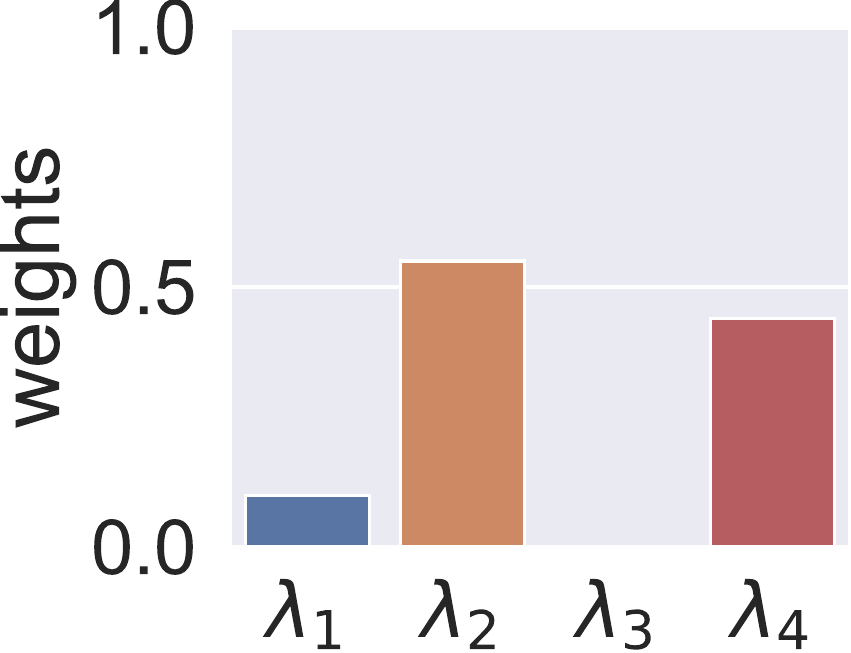}}
            }
        \end{overpic}
    \end{subfigure}%
    \hfill
    \begin{subfigure}[c]{0.248\linewidth}
        \begin{overpic}[width=\linewidth]{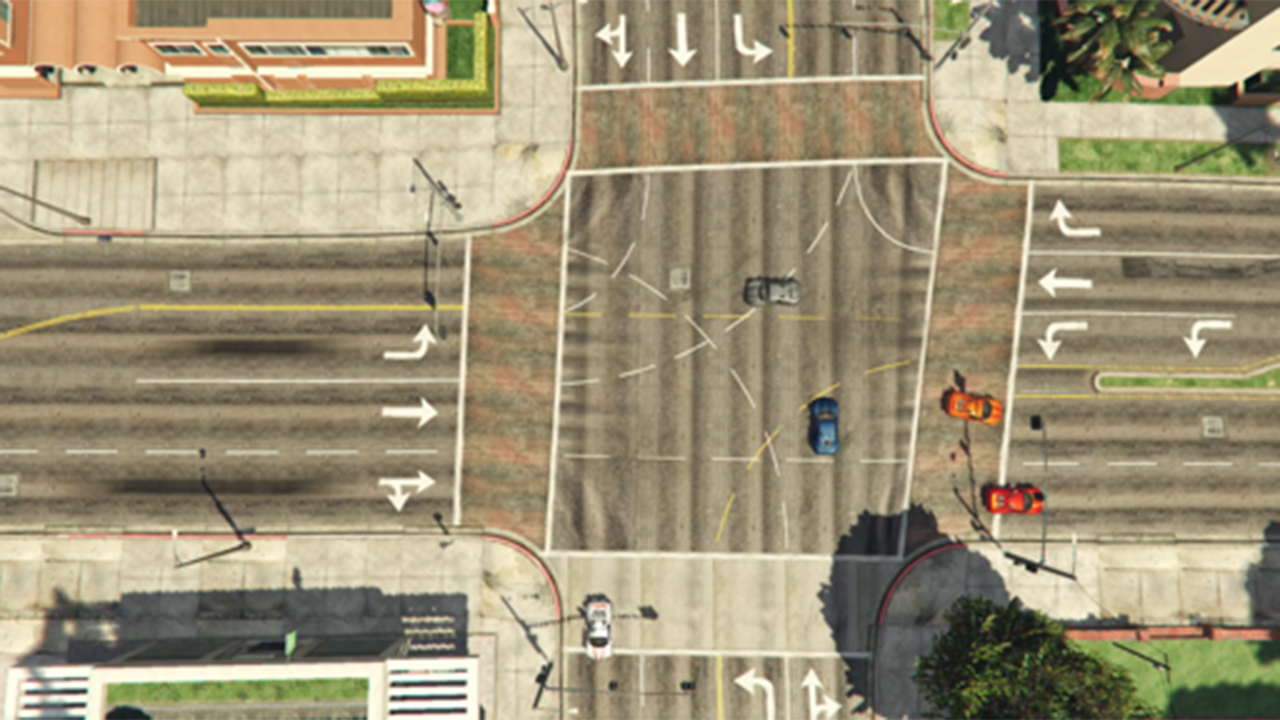}
            \put(1,23){\color{white}\linethickness{0.5mm}%
                \frame{\includegraphics[width=0.42\linewidth]{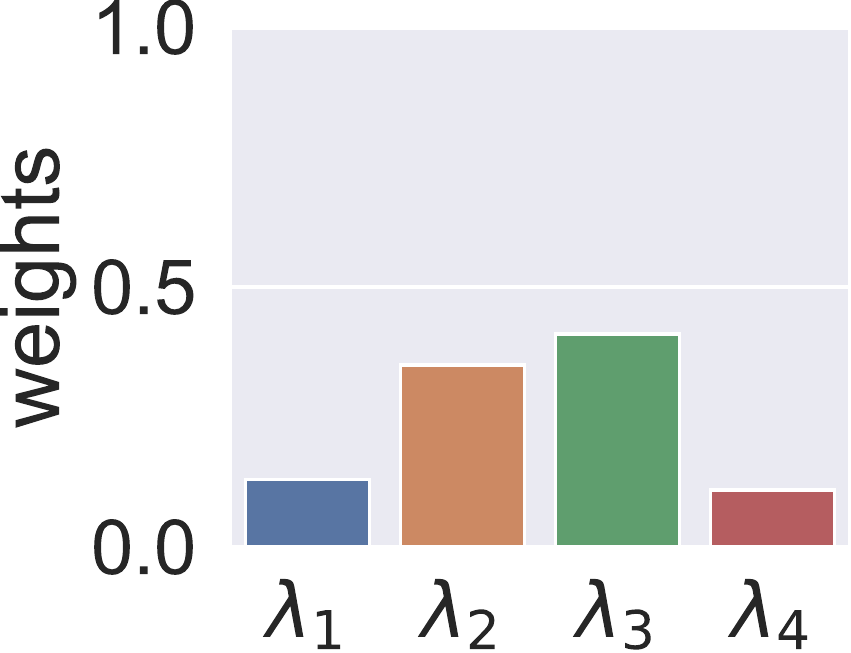}}
            }
        \end{overpic}
    \end{subfigure}%
    \caption{\textbf{Mixture weights of congestion patterns (\ac{gmm}) learned in the driving scenarios.} Top: the mixture weight distribution during overtaking. Bottom: the mixture weight distribution in a crowded intersection. Note that the vehicle number $n$ does not have to match the number of Gaussian components $M_\mathcal{Q}$.}
    \label{fig:quali_congestion}
\end{figure*}

\begin{figure*}[t!]
    \centering
    \begin{subfigure}[c]{0.248\linewidth}
        \includegraphics[width=\linewidth]{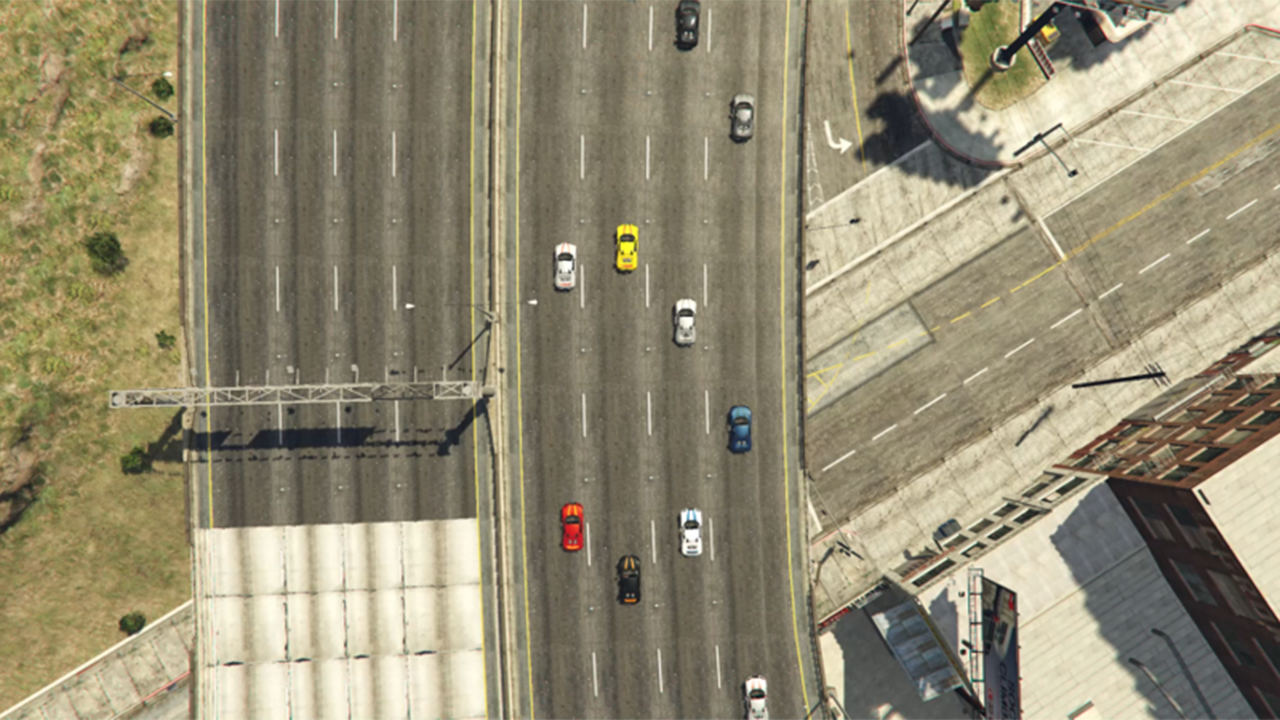}
    \end{subfigure}%
    \hfill
    \begin{subfigure}[c]{0.248\linewidth}
        \includegraphics[width=\linewidth]{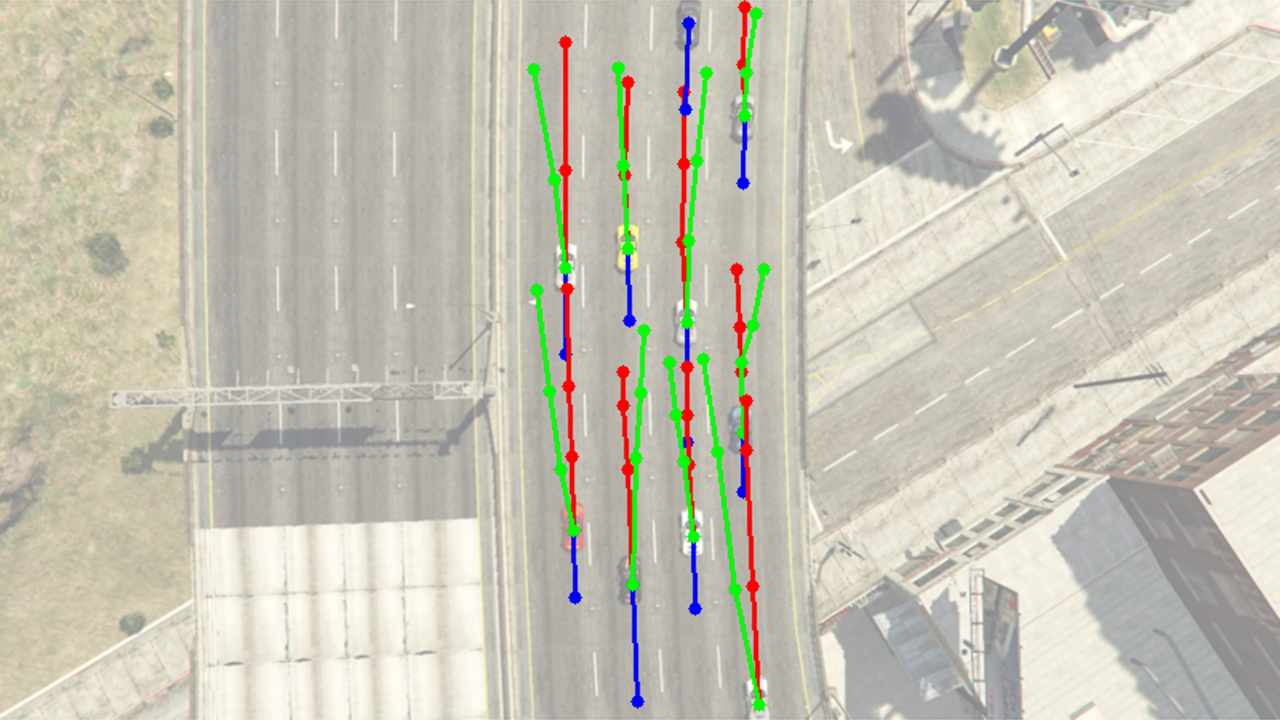}
    \end{subfigure}%
    \hfill
    \begin{subfigure}[c]{0.248\linewidth}
        \includegraphics[width=\linewidth]{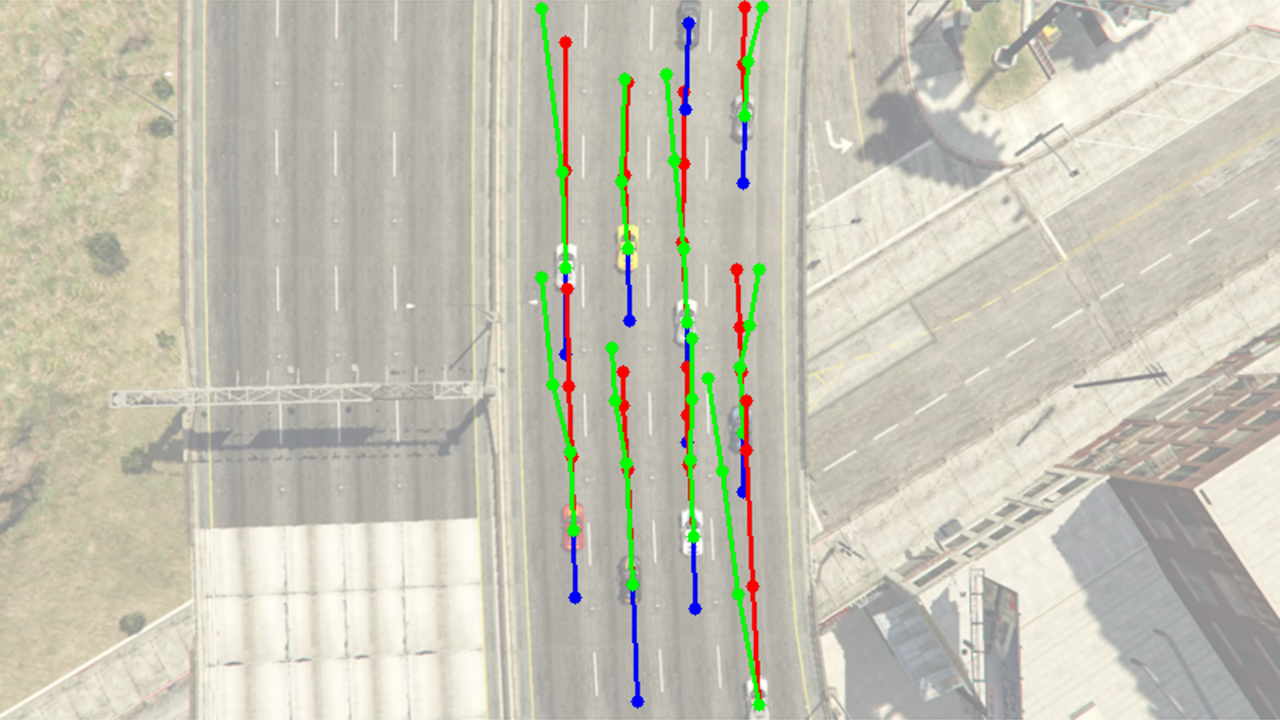}
    \end{subfigure}%
    \hfill
    \begin{subfigure}[c]{0.248\linewidth}
        \includegraphics[width=\linewidth]{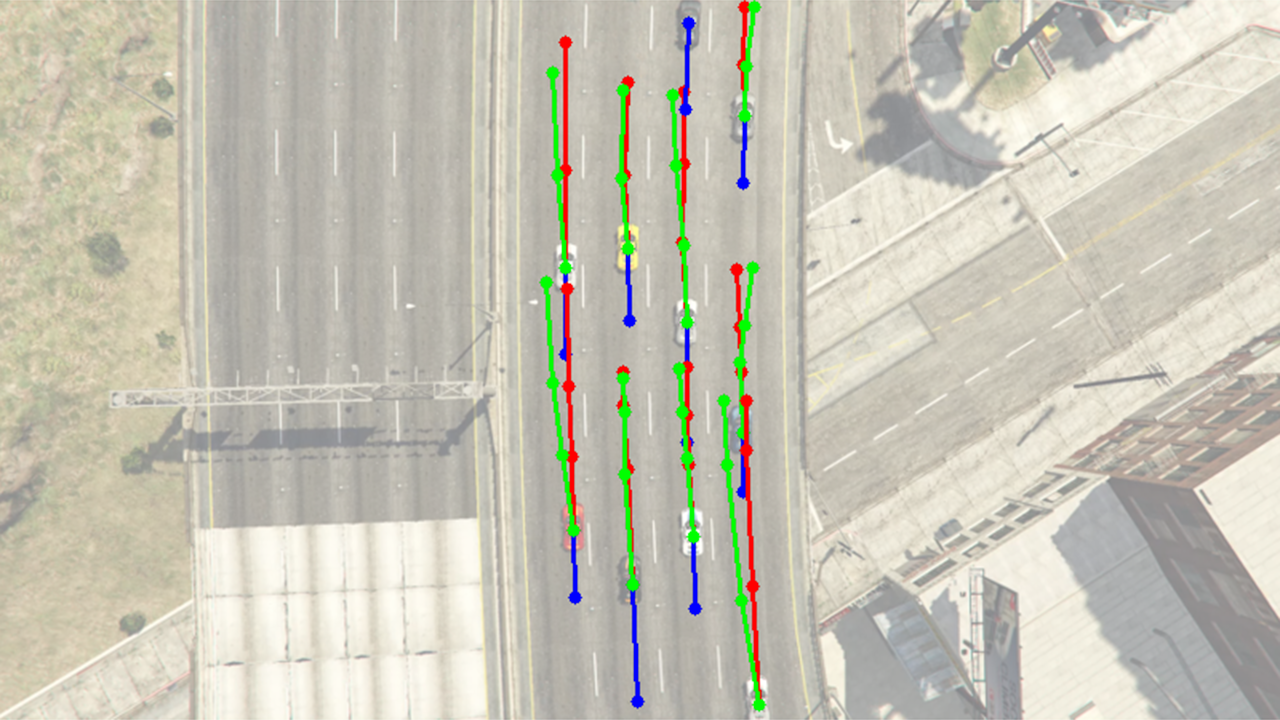}
    \end{subfigure} %
    \\\vspace{1pt}
    \begin{subfigure}[c]{0.248\linewidth}
        \includegraphics[width=\linewidth]{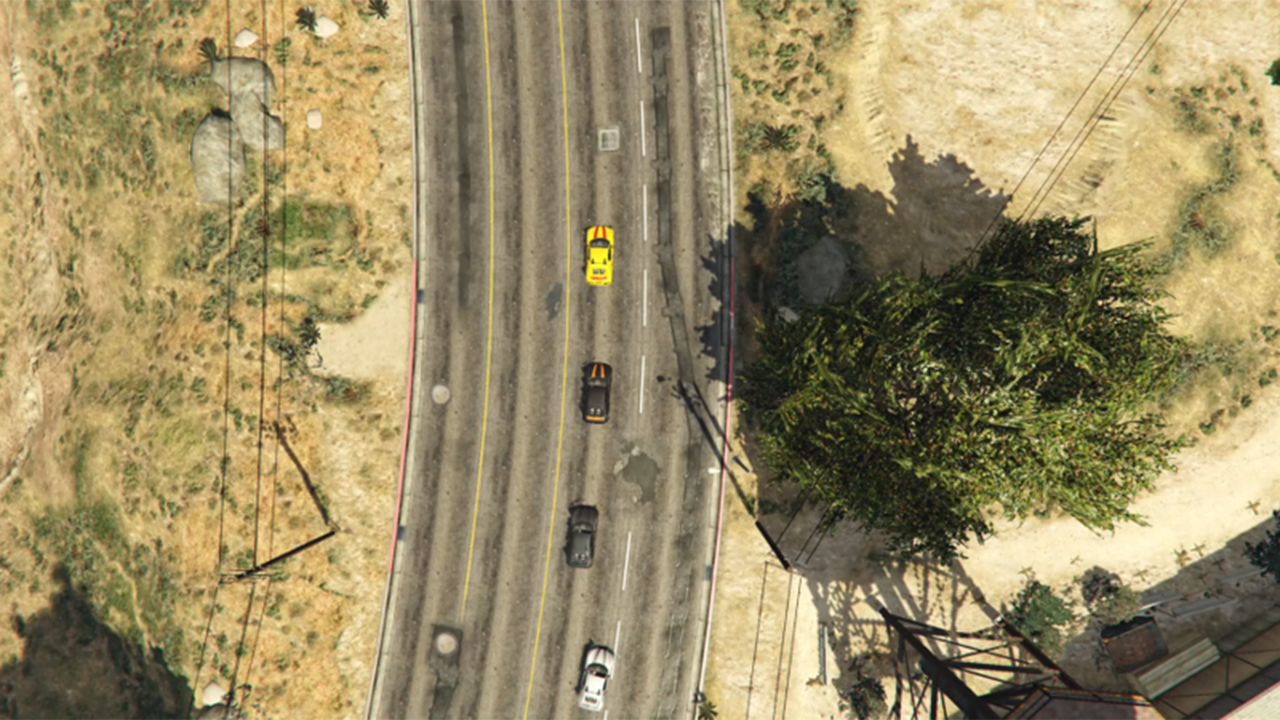}
    \end{subfigure}%
    \hfill
    \begin{subfigure}[c]{0.248\linewidth}
        \includegraphics[width=\linewidth]{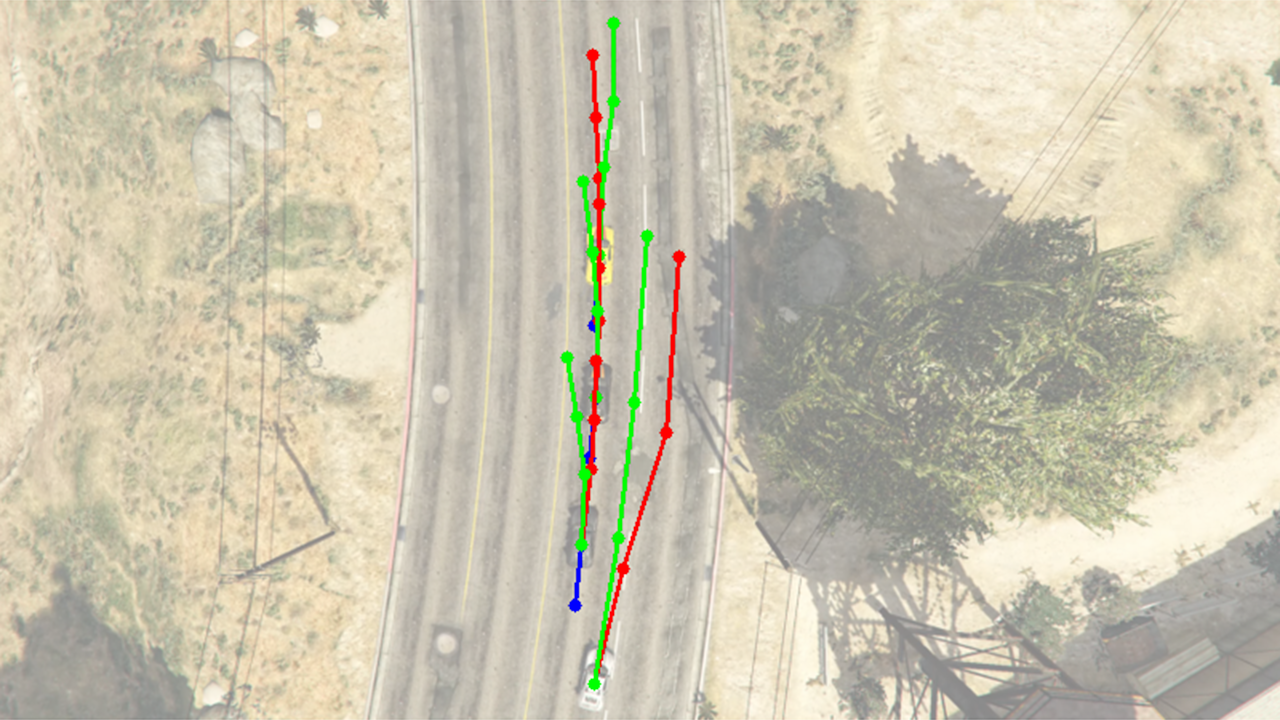}
    \end{subfigure}%
    \hfill
    \begin{subfigure}[c]{0.248\linewidth}
        \includegraphics[width=\linewidth]{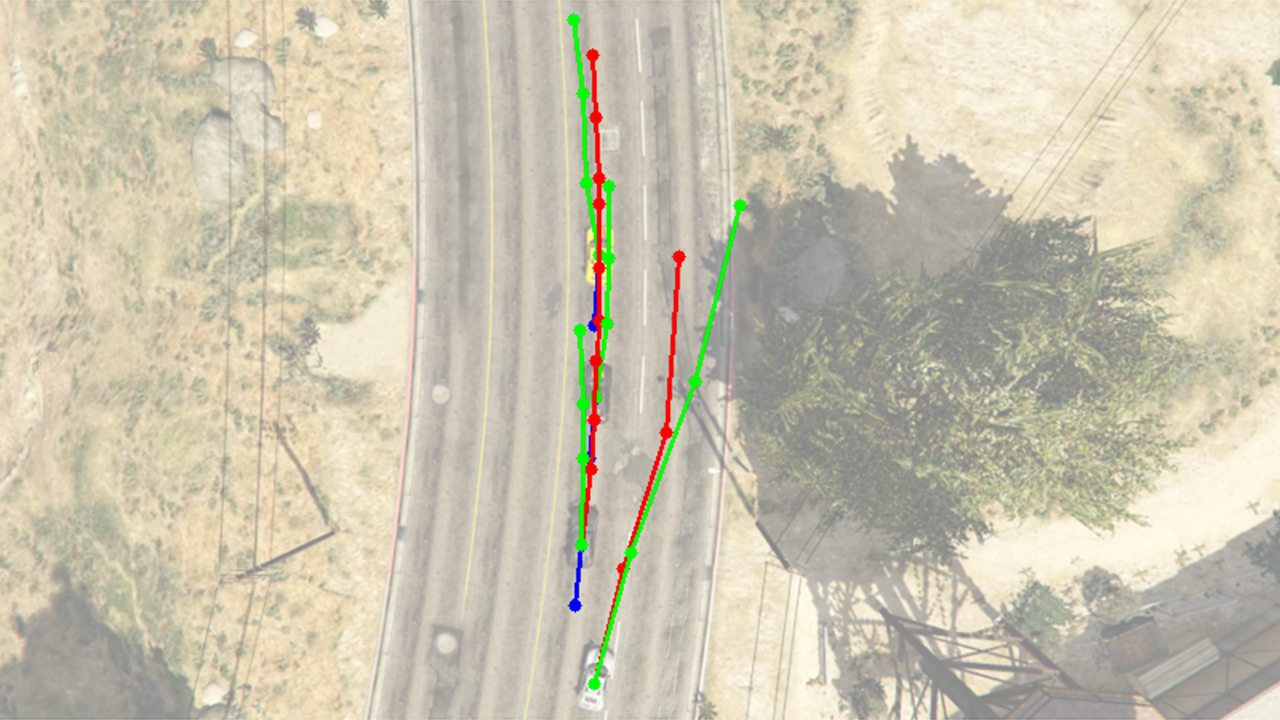}
    \end{subfigure}%
    \hfill
    \begin{subfigure}[c]{0.248\linewidth}
        \includegraphics[width=\linewidth]{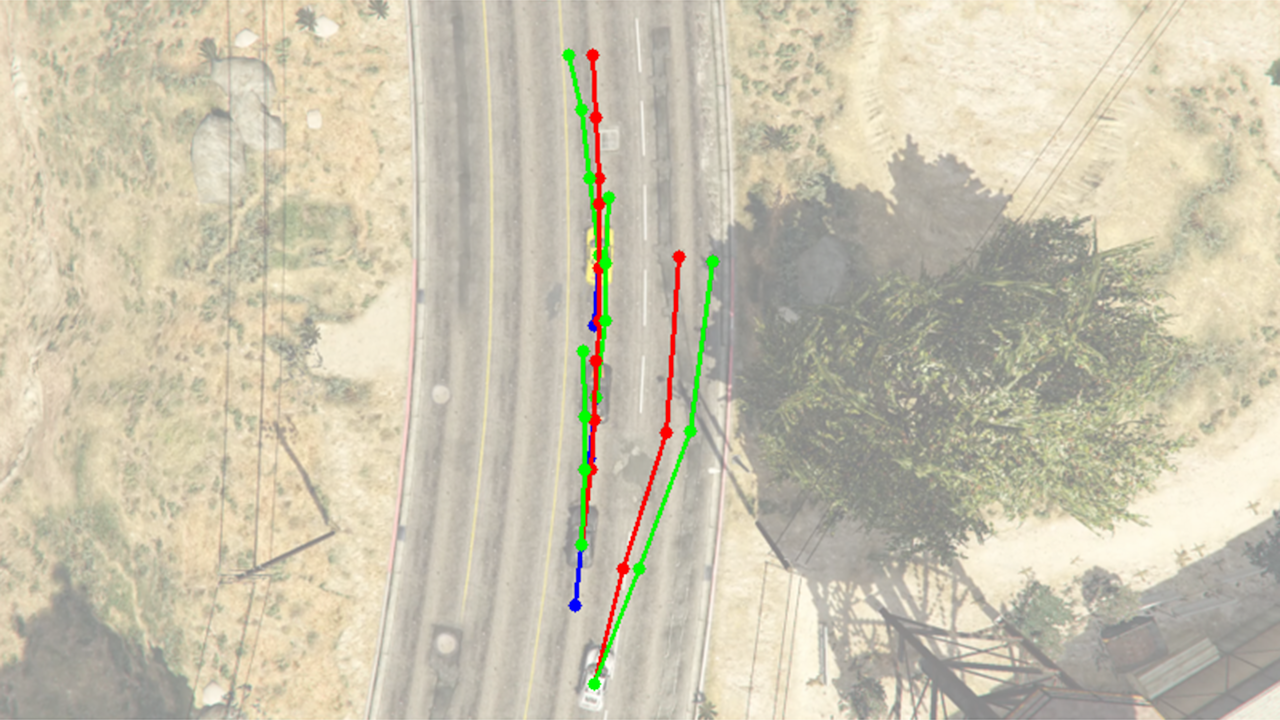}
    \end{subfigure}%
    \\\vspace{1pt}
    \begin{subfigure}[c]{0.248\linewidth}
        \includegraphics[width=\linewidth]{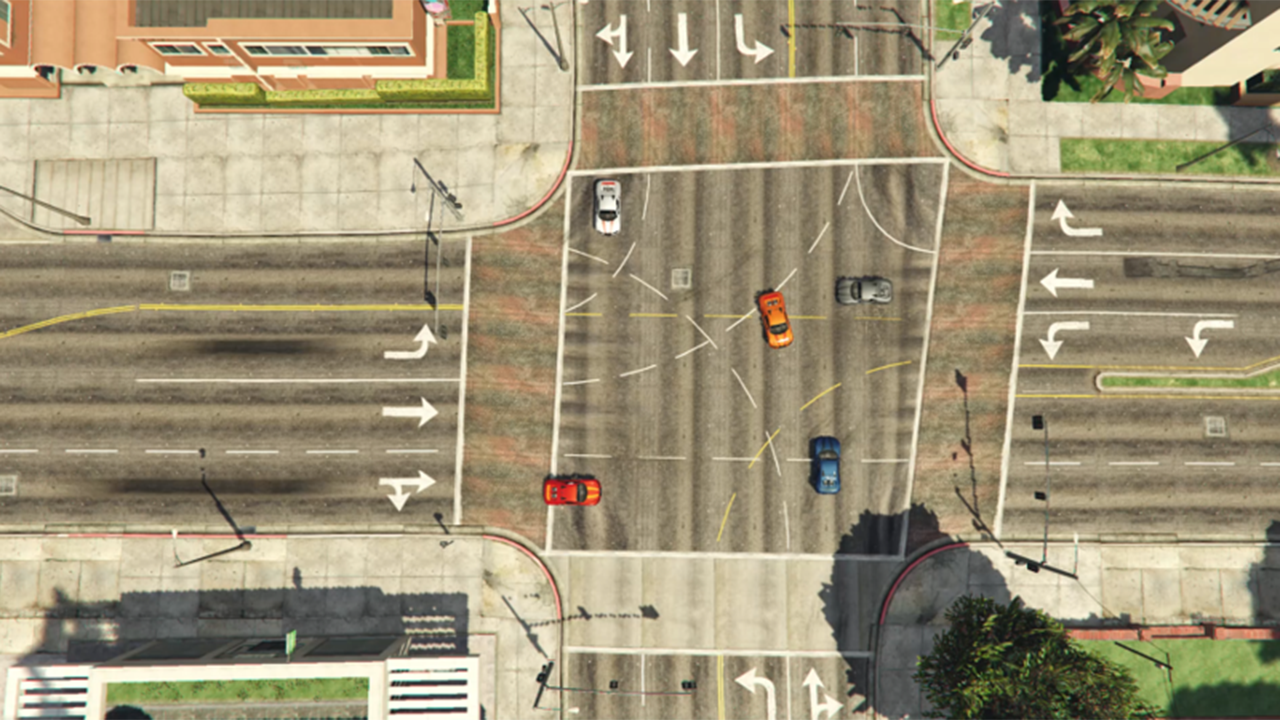}
    \end{subfigure}%
    \hfill
    \begin{subfigure}[c]{0.248\linewidth}
        \includegraphics[width=\linewidth]{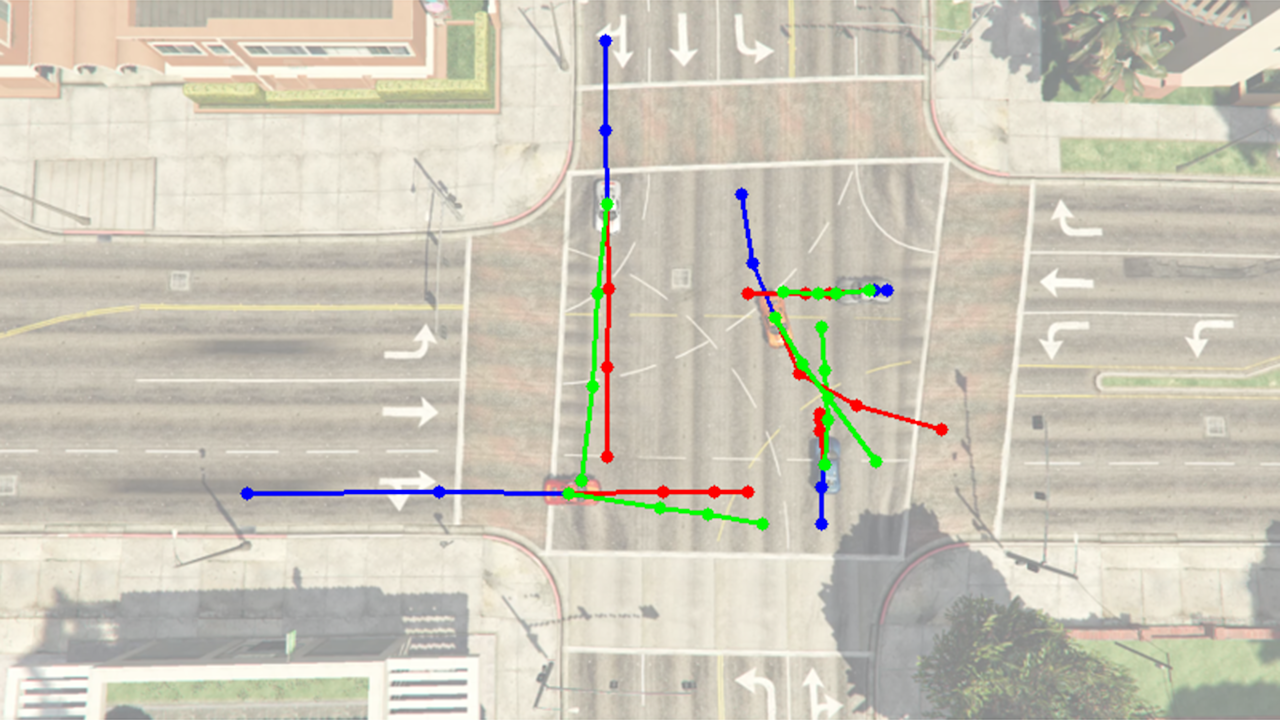}
    \end{subfigure}%
    \hfill
    \begin{subfigure}[c]{0.248\linewidth}
        \includegraphics[width=\linewidth]{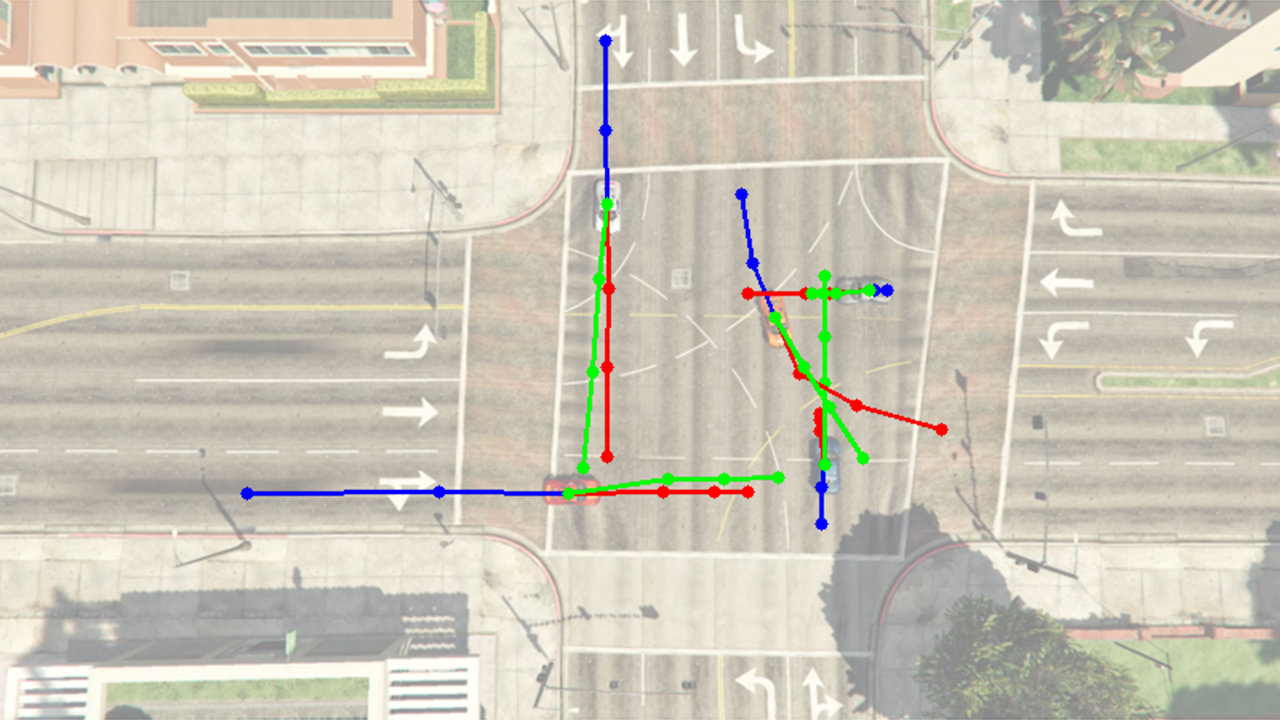}
    \end{subfigure}%
    \hfill
    \begin{subfigure}[c]{0.248\linewidth}
        \includegraphics[width=\linewidth]{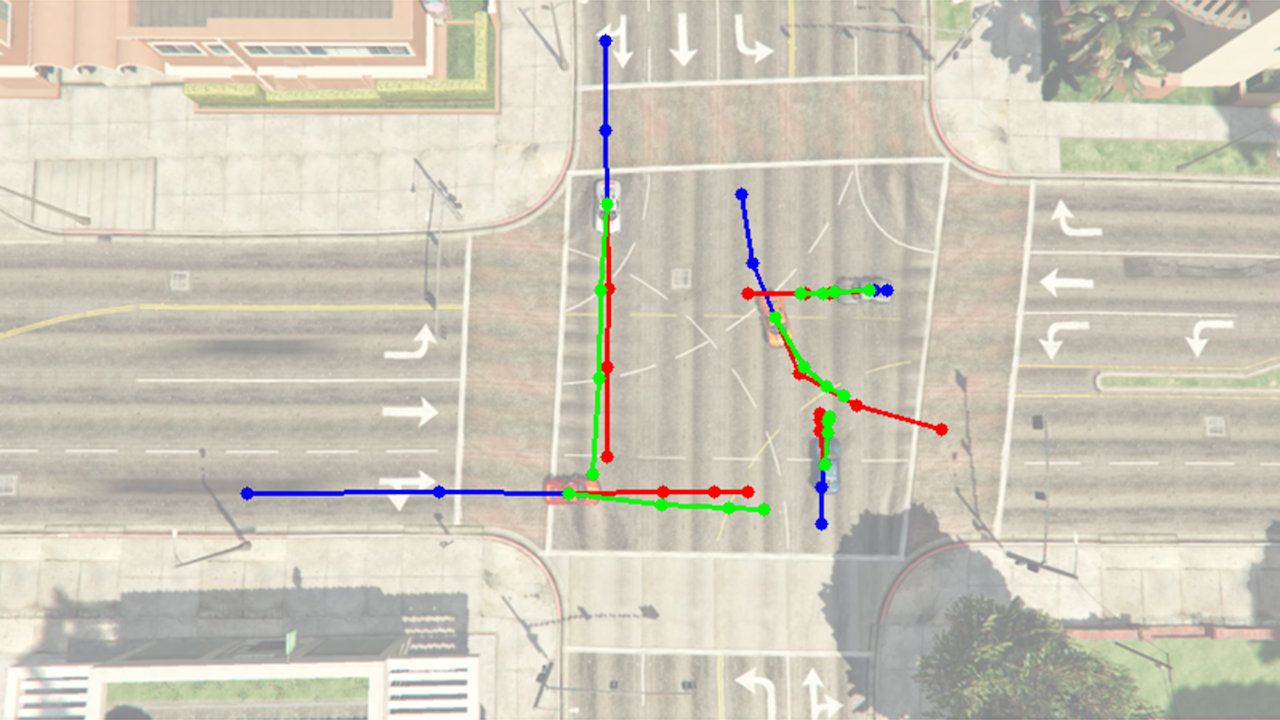}
    \end{subfigure}%
    \\\vspace{1pt}
    \begin{subfigure}[c]{0.248\linewidth}
        \includegraphics[width=\linewidth]{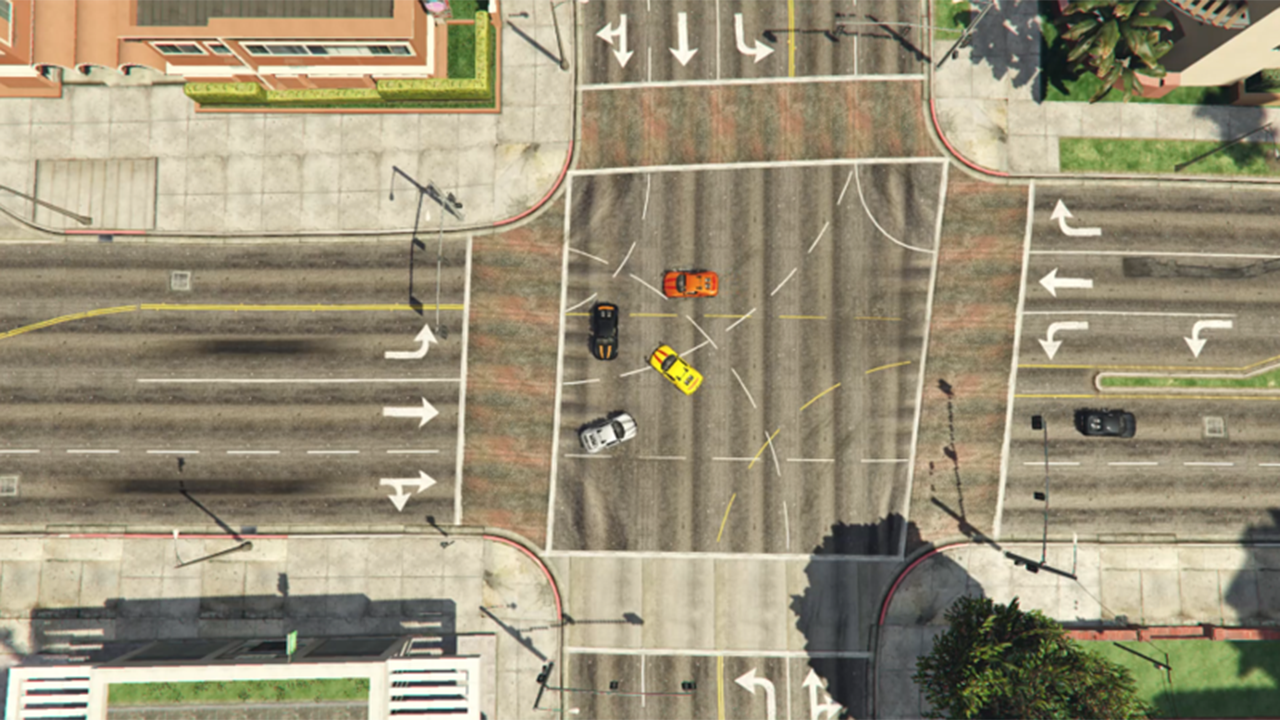}
        \caption{Scenarios}
    \end{subfigure}%
    \hfill
    \begin{subfigure}[c]{0.248\linewidth}
        \includegraphics[width=\linewidth]{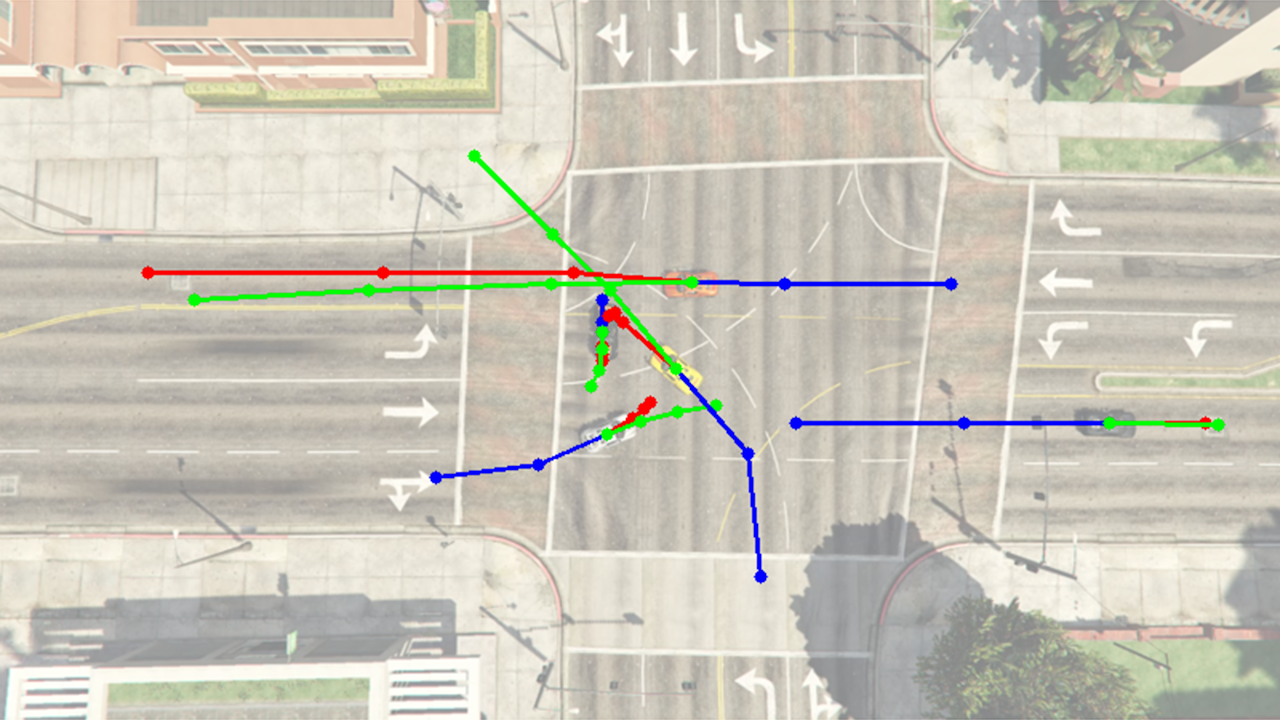}
        \caption{S-GAN}
    \end{subfigure}%
    \hfill
    \begin{subfigure}[c]{0.248\linewidth}
        \includegraphics[width=\linewidth]{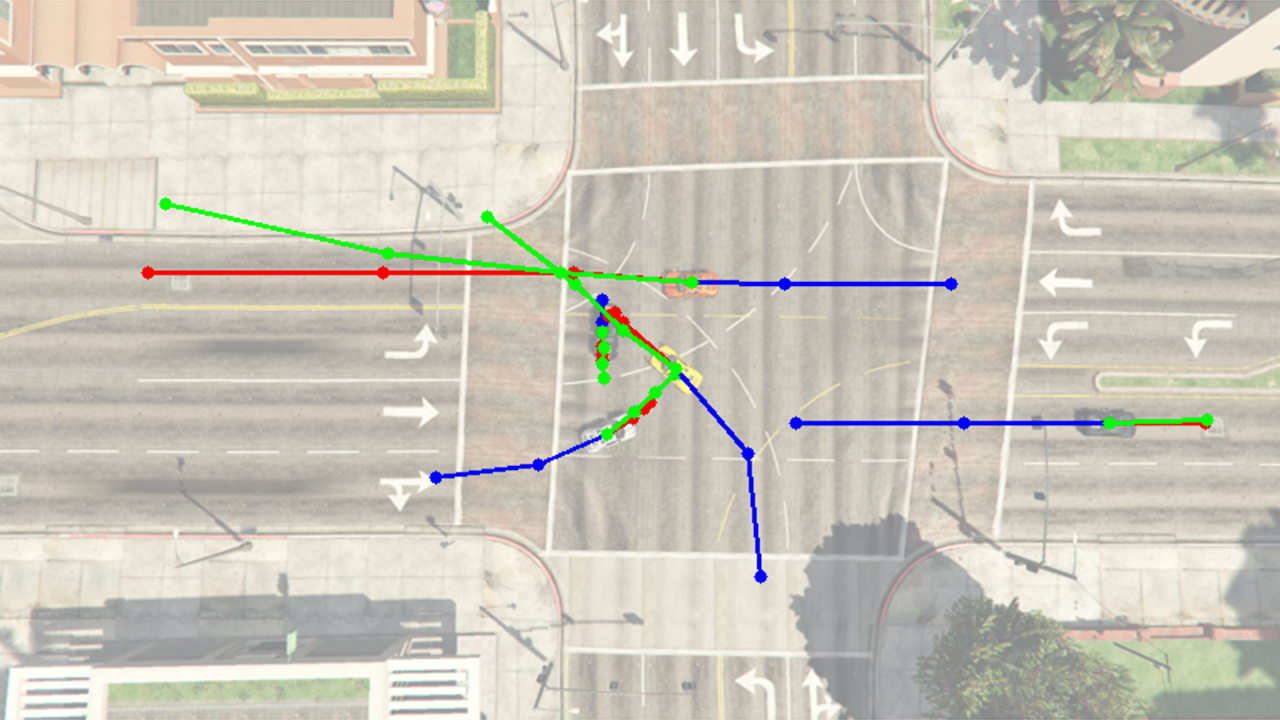}
        \caption{CS-LSTM}
    \end{subfigure}%
    \hfill
    \begin{subfigure}[c]{0.248\linewidth}
        \includegraphics[width=\linewidth]{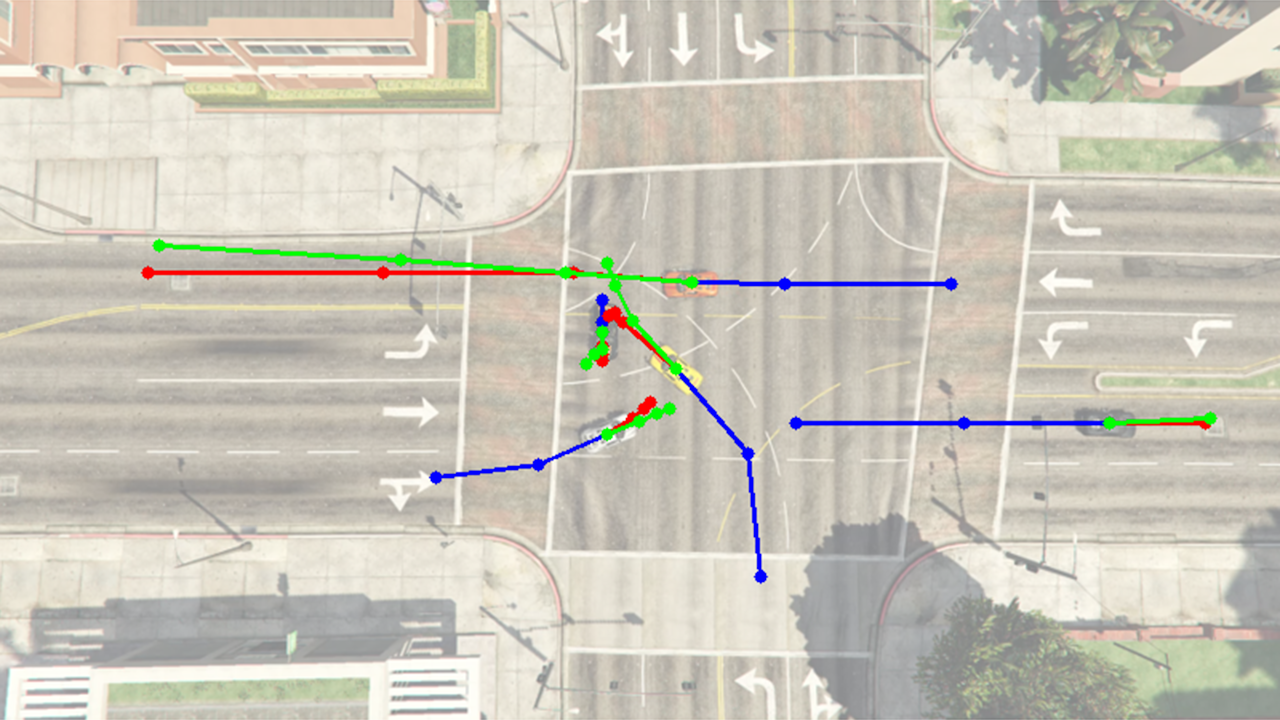}
        \caption{CF-LSTM}
    \end{subfigure}%
    \caption{\textbf{Qualitative results of trajectory prediction in four types of scenarios from the proposed CF-LSTM and two baselines.} Blue: observed trajectories. Red: ground-truth future trajectories. Green: predicted future trajectories.}
    \label{fig:quali_traj}
\end{figure*}

\paragraph*{Congestion Patterns}

We qualitatively examine our learned congestion patterns on the GTA dataset. In \cref{fig:quali_congestion}, we show the distribution $\lambda_i$ ($M_\mathcal{Q} = 4$) of the learned \ac{gmm} $\mathcal{Q}(o)$ on two different driving scenarios. The top row shows a series of driving behaviors involving lane changing and overtaking. When the overtaking occurs, the significance of the second component becomes rather evident, compared to the relatively uniform distributions at the start and end of the series. Such a distributional shift is reflected in the bottom two rows as well. In this scenario, multiple vehicles are driving into the intersection and must yield to each other to avoid the collision. As agents are getting closer to each other, making collisions more likely, some mixture weights are firing compared to the start and end frames. Taken together, these observational results verify that the learned congestion patterns can indeed reflect the contextual semantics of congestion. See ablation study for discussions on selecting the component numbers $M_\mathcal{Q}$ and $M_\mathcal{P}$.

\paragraph*{Trajectory Predictions}

We show a set of qualitative results on trajectory prediction in \cref{fig:quali_traj}. Specifically, we compare the proposed CF-LSTM with CS-LSTM~\cite{deo2018convolutional} and S-GAN~\cite{gupta2018social} on four types of driving scenarios in the GTA dataset. As shown in the first row, our method generates more accurate trajectory points for the time interval than other baselines. For the local driving scenario with overtaking shown in the second row, our method successfully captures the overtaking vehicle's behavior. The predicted trajectory keeps a relatively safe distance from other vehicles during the lane change. Other methods either fail to keep a safe driving distance or diverge from the planned trajectory. For the intersection driving scenario in the third row, our method shows the vehicles' tendency to yield to avoid the collision when they get closer, demonstrating safety awareness from the contextual cues of congestion with more reasonable vehicle driving behaviors. The last row presents the case where one vehicle is aggressively driving through the intersection and eventually crashes into another. CS-LSTM and S-GAN are unaware of this dangerous situation, while our method shows the evident deceleration and yielding behavior. These qualitative analyses indicate the efficacy of the proposed model on safety-critical driving scenarios.

\paragraph*{Ablation Study}

We also conduct an ablation study to verify the efficiency of the proposed approach on collision-free trajectory prediction. Specifically, we show that our approach is compatible with other encoder-pooling-decoder architectures. By swapping the current architecture design to S-LSTM, we further improve S-LSTM's performance on both datasets. We compare whether directly enforcing the student model to match the latent congestion features could be better than distributional modeling and find that building a multi-modal distribution on congestion patterns can significantly improve performance. We hypothesize that this is because the \ac{gmm} can account for the various modes in congestion patterns. Finally, we search for the hyperparameter on the number of mixtures. We notice that, coherent to the assumption on congestion pattern learning, the model achieves the best performance when the hidden mixture number equals that of the ground-truth. Please refer to the supplementary video for details of the ablation study.

\setstretch{0.97}

\section{Conclusions}\label{sec:conclusion}

In this work, we study the problem of multi-agent trajectory prediction. We propose to explicitly learn congestion patterns as contextual cues and decouple the ``Sense-Learn-Reason-Predict'' framework into a teacher-student process. We formulate an optimization problem to bridge the connection between the congestion patterns and the learning objective of collision-free trajectory prediction. In experiments, we show that the proposed model is able to achieve the best performance on collision-free trajectory prediction on a synthetic dataset designed for collision avoidance evaluation while remaining competitive on regular trajectory prediction on the NGSIM US-101 highway dataset.

\bibliographystyle{ieeetr}
\bibliography{IEEEfull}

\end{document}